\documentclass[runningheads]{llncs}

\usepackage[mobile]{eccv}

\usepackage{eccvabbrv}

\usepackage{graphicx}
\graphicspath{{figures/}}  
\usepackage{tablefootnote}
\usepackage{booktabs}
\usepackage[accsupp]{axessibility} 
\usepackage[accsupp]{axessibility}  

\usepackage{xcolor}
\newcommand{\tokm}[1]{\textcolor{blue}{\texttt{#1}}}     
\newcommand{\toka}[1]{\textcolor{teal}{\texttt{#1}}}     
\newcommand{\toke}[1]{\textcolor{magenta}{\texttt{#1}}}  
\usepackage{tcolorbox}
\tcbuselibrary{breakable}
\usepackage{xltabular}
\usepackage{array}
\usepackage{makecell}
\usepackage{enumitem}
\newcolumntype{Y}{>{\centering\arraybackslash}X}
\usepackage{longtable}
\usepackage{orcidlink}
\usepackage{multirow}
\usepackage{hyperref}

\begin{document}

\title{ReactMotion: Generating Reactive Listener Motions from Speaker Utterance} 

\titlerunning{Abbreviated paper title}

\author{Cheng Luo\inst{1}$^\dag$\ \and Bizhu Wu\inst{2,4,5}$^\dag$\ \and Bing Li\inst{1}$^*$ \and Jianfeng Ren\inst{4}, Ruibin Bai\inst{4}, Rong Qu\inst{5} \and Linlin Shen\inst{2,3}$^*$ \and Bernard Ghanem\inst{1}  
}

\authorrunning{C. Luo et al}

\institute{King Abdullah University of Science and Technology \and
Computer Vision Institute, School of Artificial Intelligence, Shenzhen University
\and
Guangdong Provincial Key Laboratory of Intelligent Information Processing, Shenzhen University
\and
School of Computer Science, University of Nottingham Ningbo China \and
School of Computer Science, University of Nottingham, United Kingdom
\\
\textcolor{blue}{Project page: \url{https://reactmotion.github.io} }
}



\maketitle

\begingroup
\renewcommand\thefootnote{}
\footnotetext{${}^\dag$ Equal contribution. ${}^*$ Corresponding authors.}
\endgroup

\begin{abstract}
In this paper, we introduce a new task, Reactive Listener Motion Generation from Speaker Utterance, which aims to generate naturalistic listener body motions that appropriately respond to a speaker’s utterance. However, modeling such nonverbal listener behaviors remains underexplored and challenging due to the inherently non-deterministic nature of human reactions.
To facilitate this task, we present ReactMotionNet, a large-scale dataset that pairs speaker utterances with multiple candidate listener motions annotated with varying degrees of appropriateness. This dataset design explicitly captures the one-to-many nature of listener behavior and provides supervision beyond a single ground-truth motion. Building on this dataset design, we develop preference-oriented evaluation protocols tailored to evaluate reactive appropriateness, where conventional motion metrics focusing on input–motion alignment ignore. We further propose ReactMotion, a unified generative framework that jointly models text, audio, emotion, and motion, and is trained with preference-based objectives to encourage both appropriate and diverse listener responses. Extensive experiments show that ReactMotion outperforms retrieval baselines and cascaded LLM-based pipelines, generating more natural, diverse, and appropriate listener motions. 

\end{abstract}

\section{Introduction}
\label{sec:intro}

\begin{figure}[t]
  \centering
  \includegraphics[width=\linewidth]{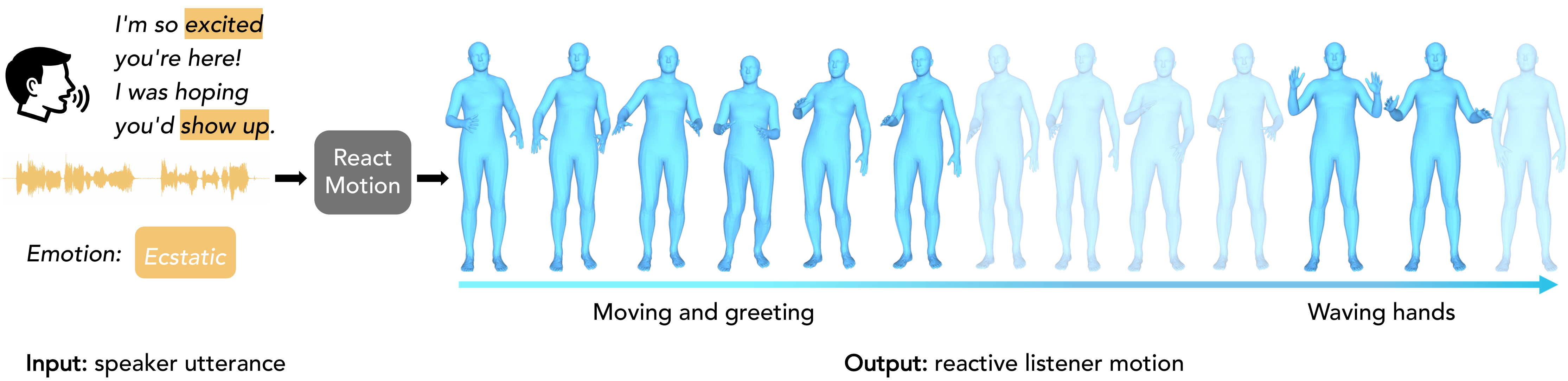}
  \setlength{\abovecaptionskip}{-5pt} 
  \caption{\textbf{Illustration} of the proposed new task: Reactive Listener Motion Generation from Speech Utterance. Given a speaker’s utterance, \ie, transcript and/or audio (optionally supplemented with emotion), a generative model such as our \textsc{ReactMotion} generates a corresponding responsive body-motion sequence for the listener.}
  \label{fig:intro-reactmotion}
\end{figure}

Modeling dyadic human communication is crucial for virtual agents~\cite{kim2023avatar}, digital humans~\cite{zhu2025infp, luoomniresponse}, and social robots~\cite{spaccatini2023new}. 
While prior work has advanced speech-to-speech dialogue~\cite{defossez2024moshi}, language-based interfaces~\cite{hurst2024gpt, achiam2023gpt}, and listener facial reactions~\cite{ng2022learning, song2024react}, reactive \emph{listener body motions} remain largely overlooked despite being central to face-to-face interaction.
Listeners often convey engagement and understanding through posture and subtle gestures, and generating such feedback is important for natural dyadic communication.

We introduce a new task, \emph{Reactive Listener Motion Generation from Speech Utterance}, which aims to generate naturalistic listener body motions that appropriately respond to a speaker’s utterance given its audio and/or transcript.
Unlike text-to-motion~\cite{CLoSD, AMD, zhang2023generating, tevet2023mdm, petrovich24stmc} or audio-driven motion generation~\cite{xu2025mospa} that primarily realize the input content, our setting models conversational reactions where speaker cues are indirect and the output is inherently one-to-many.

This task poses three challenges. (i) The same utterance can elicit multiple valid listener reactions~\cite{song2024react,ng2022learning}. Such non-deterministic listener behaviour poses a significant challenge for modeling the listener’s motion responses. (ii) There is no publicly available large-scale dataset with \emph{multiple} listener-reactive \emph{body motions} per utterance, to the best of our knowledge. (iii) Reactive appropriateness is difficult to evaluate. Metrics based on a single ground truth or motion diversity are insufficient to measure the appropriateness of a listener’s reaction.

To address these challenges, we introduce \textbf{ReactMotionNet}, a curated dataset with 151,328 (speaker utterance, listener motion) pairs.
\textbf{Unlike prior motion datasets that typically provide a single target per condition}, we associate each utterance with \textit{multiple} candidate reactions and annotate them into three preference tiers, \textit{Gold}, \textit{Silver}, and \textit{Negative}.
This tiered design captures one-to-many ambiguity and enables \textbf{preference-style supervision and evaluation}~\cite{chiang2024chatbot, zheng2023judging, christiano2017deep}.
Moreover,  we propose a scalable pipeline that re-purposes existing motion data into dyadic speaker-listener pairs for dataset construction, which \textbf{avoids relying on expensive speaker--listener motion capture}.

To evaluate reactive appropriateness, we introduce a \textbf{tier-aware ranking protocol}.
We train a multimodal judge network to score and rank candidate reactions under the same speaker input and report win rates against the Gold, Silver, or Negative tiers.
\textbf{This relative evaluation goes beyond single-reference similarity} and better reflects that multiple reactions can be appropriate for the same utterance.
Finally, we propose \textbf{ReactMotion}, a unified generative framework that jointly models speaker transcript, emotion, and audio to generate listener motions.
We leverage the tiered annotations with preference-based objectives that learn from \emph{relative} comparisons within each utterance group for the training.

\paragraph{Contributions.}
(i) To the best of our knowledge, we introduce the first task of reactive listener \emph{body} motion generation from speaker speech in dyadic interaction.
(ii) We present \textbf{ReactMotionNet}, a new dataset with multi-tier (Gold/Silver/Negative) reactive listener motions and a tier-aware evaluation protocol for reactive appropriateness, enabling research on nonverbal listener response behavior.
(iii) We propose \textbf{ReactMotion}, a unified multimodal generative model that processes multiple speaker cues and generates high-quality listener body motions in response to the speaker.

\section{Related Work}
\label{sec:related_work}

\textbf{Human Motion Generation.}
Human motion generation can be conditioned on diverse modalities, including text~\cite{zhang2025kinmo, liao2025shape, meng2025rethinking, wang2025stickmotion, zhang2025energymogen, Chen_2025_CVPR, lu2025scamo, HGM3, pinyoanuntapong2024controlmm, MotionStreamer}, 
action classes~\cite{petrovich2021actor, tevet2022motionclip, raab2023modi}, and
audio signals such as music~\cite{li2022danceformer, li2024exploring, li2025lodge++, yang2025lagrangian} or speech~\cite{xu2025combo, li2023audio2gestures, liu2024towards}). 
Among these, text- and audio-driven motion generation are most related to our setting. 
Text-based approaches generate motions from explicit action descriptions~\cite{parco, huang2024como, guo2024momask, wang2023fgt2m, fgt2m++, zhang2024motiongpt, petrovich2022temos, kim2023flame, barquero2024flowmdm, chen2023mld, zhang2024motiondiffuse}, 
while audio-driven methods synthesize gestures aligned with temporally synchronized acoustic signals~\cite{mughal2024convofusion, chen2024enabling, zhang2025semtalk}.
Representative modeling paradigms include transformer-based latent models (\textit{e.g.},~\cite{petrovich2021actor, zhang2025echomask, liu2024emage}), 
discrete motion tokenization with autoregressive modeling (\textit{e.g.},~\cite{zhang2023generating, yi2023generating, ao2022rhythmic, chen2025language}), 
and diffusion-based frameworks (\textit{e.g.},~\cite{tevet2023mdm, alexanderson2023listen, he2024co, liu2025gesturelsm}).

Beyond single-person generation, recent works~\cite{liang2023intergen, wang2024intercontrol, mughal2024convofusion, ho2025interact, ng2024audio, sun2025beyond} extend motion synthesis to multi-person scenarios. 
These approaches typically generate multi-person motions by conditioning on \textit{explicit} textual descriptions of joint actions or on the audio streams of both individuals. 
In contrast, our problem setting differs in that the target motion is not directly specified by explicit action instructions or synchronized signals. 
Instead, the model must infer the \textit{implicit} interaction intention from the speaker’s utterance, including transcript, audio, and emotion cues, and produce a socially appropriate reactive motion for the listener. 
This requires reasoning over cross-speaker dynamics rather than direct condition-to-motion mapping.

\noindent\textbf{Human Reaction Generation.} Human reaction generation is crucial for AI interaction systems. 
Spoken language modeling has progressed from cascaded ASR $\rightarrow$ LLM $\rightarrow$ TTS pipelines to end-to-end and full-duplex speech-to-speech models~\cite{rubenstein2023audiopalm, zhang2023speechgpt, defossez2024moshi, veluri2024syncllm}, while facial reaction generation has advanced from conditional GANs~\cite{huang2017dyadgan} to uncertainty-aware and diffusion-based methods~\cite{ng2022learning, zhou2022rlhg, luo2024reactface, luo2025reactdiff, song2024react}. 
Audio-visual face-to-face dialogue modeling has been explored~\cite{park2024f2f, ng2022learning, zhou2022rlhg, chu2025unils}.

In 3D human body modeling, most methods synthesize reactor motion conditioned on actor motion~\cite{chopin2023interaction,ghosh2024remos,liu2023interactive,liu2024physreaction,xu2024regennet}. 
For instance, InterFormer~\cite{chopin2023interaction} uses temporal-spatial attention in Transformers, and ReGenNet~\cite{xu2024regennet} and ReMoS~\cite{ghosh2024remos} employ diffusion models for full-body motion. 
Recently, HERO~\cite{yu2025hero} generates 3D reactive motion directly from RGB videos, incorporating the actor’s facial expressions to capture emotional cues. Differently, our method generates 3D reactor motion from the speaker’s utterance, which includes transcript, audio, and optional emotion annotations. Transcript provides a lightweight, user-friendly modality, audio offers rich vocal cues, and emotion labels explicitly indicate mood, facilitating more effective interaction modeling.

\noindent\textbf{3D Human Body Interaction Datasets.}
Recent datasets have facilitated research on multi-person dynamics and interaction-aware 3D motion. 
Several works~\cite{guo2022multi, hu2013efficient, liang2023intergen, xu2024interx, yin2023hi4d} provide paired human motions, modeling interaction as symmetric kinematic coupling, where one participant’s motion is predicted from the other’s. While effective for spatial coordination, this ignores linguistic and affective signals that drive conversation.

Other datasets~\cite{yu2025hero, khirodkar2023egohumans, khirodkar2024harmony4d, ko2021air, ng2020you2me, ryoo2013first, ryoo2015robot} supply silent RGB videos with 3D reactive motions, offering richer context but still lacking speech semantics and emotional cues, which are central to communicative intent.
Some datasets~\cite{ho2025interact, lee2019talking, ng2024audio, sun2025beyond} include both audio and motion for human interactions, but the movements of their motions primarily focus on the upper body, such as arms, and are limited to one-to-one speaker-listener pairs.

In contrast, our dataset provides a one-to-many mapping between speaker utterances and listener reactive motions. 
Each utterance has multiple responses labeled \textit{gold}, \textit{silver}, and \textit{neg} for appropriate, partially appropriate, and irrelevant reactions, making it better suited for practical applications. Plus, motions are more dynamic, such as jumping, enabling more diverse body reactions.

\section{Task Definition}
\label{sec:task_definition}
In this paper, we study \emph{Reactive Listener Motion Generation} in dyadic interaction, which consists of a \emph{speaker} and a \emph{listener}. 
Given a speaker utterance $C^s$, the goal is to generate appropriate reactive body motion of the listener, denoted as $R^l$. 
Formally, the objective is to learn the conditional distribution:
\begin{equation}
p_\theta\!\left(R^l \mid C^s\right), 
\qquad
C^s \in \Big\{A^s,\; T^s,\; (A^s, T^s),\; (A^s, E^s),\; (T^s, E^s),\; (A^s, T^s, E^s)\Big\}.
\label{eq:task_def}
\end{equation}
Here, $A^s$ denotes the speaker audio, $T^s$ is the corresponding textual transcript, $E^s$ represents the speaker emotion, and $\theta$ denotes the model parameters. 
As shown in Eqn.~\ref{eq:task_def}, $C^s$ may consist of different modalities of the speaker utterance or their combinations.
At inference time, diverse listener reactions can be sampled from $p_\theta(R^l \mid C^s)$.

In contrast to conventional text-to-motion generation, the speaker utterance do not explicitly specify the target listener motion. The mapping from $C^s$ to $R^l$ is therefore inherently one-to-many, which requires the model to generate motions that are contextually appropriate while maintaining diversity.

\section{ReactMotionNet Dataset}
\label{sec:dataset}

\begin{figure}[t]
  \centering
  \includegraphics[width=\linewidth]{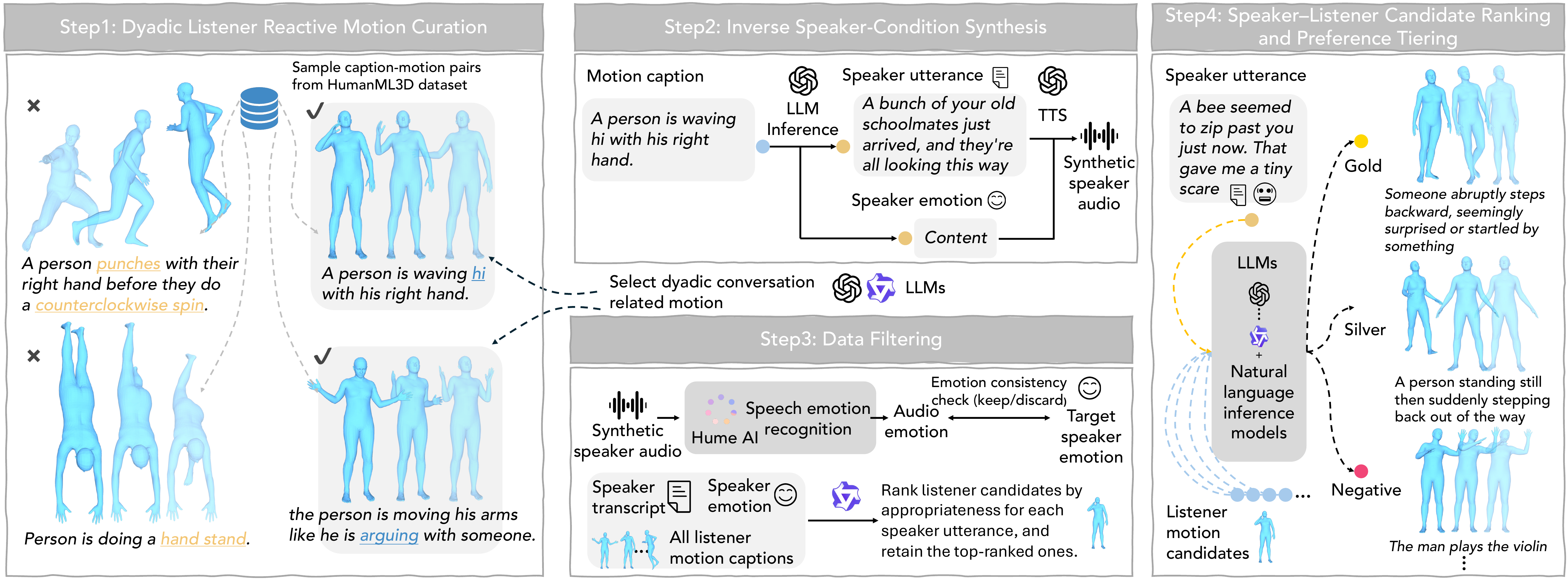}
  \setlength{\abovecaptionskip}{-5pt}
  \caption{\textbf{ReactMotionNet dataset construction.}
We curate dyadic listener motions (Step~1), synthesize speaker conditions via inverse inference and Text-to-Speech (TTS) (Step~2), filter unreliable samples (Step~3), and rank/re-tier speaker--listener pairs into \emph{gold/silver/negative} preferences (Step~4).}
  \label{fig:dataset-construction-reactmotion}
\end{figure}

To bridge the gap between existing 3D human motion interaction datasets and real-world conversational dynamics, we construct a dataset, ReactMotionNet, featuring \textit{one-to-many} speaker utterance–listener reaction mappings with \textit{graded appropriateness annotations}.
To construct this dataset, we present a novel data construction pipeline (Fig.~\ref{fig:dataset-construction-reactmotion}) that repurposes existing human motion data into speaker–listener motion–response pairs using powerful LLMs~\cite{qwen3, openai_o3mini_2025}, thereby avoiding costly data collection.

\subsection{Dataset Construction Pipeline}
\label{sec:dataset_construction}

\paragraph{Step 1: Dyadic Listener Reactive Motion Curation.}
Unlike existing audio-driven 3D human interaction datasets, which mainly focus on upper-body movements while standing still, we curate motions from the more dynamic and commonly used HumanML3D dataset~\cite{guo2022generating}. 
Leveraging the textual captions of motions, we filter out conversation-irrelevant ones (\textit{e.g.}, doing a handstand) using multiple LLM-based verifiers (e.g., ChatGPT-o1~\cite{jaech2024openai}, ChatGPT-o3 mini~\cite{openai_o3mini_2025}). 
This step results in a set of motions with reaction-like semantics, which serve as the listener’s reactive motions.

\paragraph{Step 2: Inverse Speaker-Condition Synthesis.}
For each listener motion $R^l$ from the last step, we infer multiple plausible speaker utterances that could elicit the observed reaction.
Concretely, we input the listener motion's caption into OpenAI o3‑mini~\cite{openai_o3mini_2025, singh2025openai, achiam2023gpt} to generate potential speaker transcripts $T^s$ and associated emotion labels $E^s$.
We incorporate emotion into utterance generation, as the speaker’s emotional state influences the listener’s reaction. 
For example, the same transcript, “Do whatever you want,” can lead to different responses: a supportive tone may cause the listener to jump happily in place, whereas a frustrated tone may cause the listener to walk away feeling hurt.
Given $T^s$ and $E^s$, we  synthesize the corresponding speaker audio $A^s$ using GPT-4o mini TTS~\cite{hurst2024gpt}.
These steps produce a pool of possible speaker utterances ($A^s$, $T^s$, $E^s$).

\paragraph{Step 3: Data Filtering.}
We perform a series of procedures to ensure the dataset quality.
First, for each speaker utterance, we verify whether the synthesized audio $A^s$ faithfully reflects the intended emotion $E^s$. Specifically, we apply an automatic speech emotion recognizer (\emph{i.e.,} Hume AI~
\footnote{\url{https://www.hume.ai/expression-measurement}})
to the generated audio and discard any utterance whose predicted emotion is inconsistent with its assigned emotion label.
Next, we pair each remaining speaker utterance with the caption of every listener reactive motion $R^l$ obtained in Step 1. 
We then employ Qwen (Qwen3-235B-A22B-Instruct)~\cite{qwen3} to assign a dyadic conversation appropriateness score to each speaker-utterance and listener motion caption pair. 
For each speaker utterance, we retain only the top several higher-scoring listener reactive motions, thereby removing inappropriate pairs.

\paragraph{Step 4: Speaker–Listener Candidate Ranking and Preference Tiering.}
Given a pair consisting of a speaker utterance and one of its corresponding listener reactive motions from Step 3, we use multiple agents (\emph{i.e.,}  ChatGPT-o1~\cite{jaech2024openai}, ChatGPT-o3 mini~\cite{openai_o3mini_2025}, and Qwen3-235B-A22B-Instruct~\cite{qwen3}) to evaluate the pair. 
They score it according to
(1) \emph{semantic appropriateness} (whether the reaction fits the utterance), 
and (2) \emph{conversational plausibility} (whether it sounds like a natural dyadic response).
We further use a natural language inference (NLI) model\footnote{\url{https://huggingface.co/MoritzLaurer/deberta-v3-large-zeroshot-v1.1-all-33}}
to verify whether the listener motion caption is a logically plausible inference from the speaker utterance.
We then weighted sum the agents’ scores to obtain a final score, which is used to label the pair as \emph{gold}, \emph{silver}, or \emph{negative} according to predefined thresholds.

\begin{table}[t]
\setlength{\belowcaptionskip}{5pt} 
\caption{
\textbf{Dataset statistics.}
\#Pairs is the total number of labeled speaker--listener pairs (\emph{i.e.} candidate reactions).
\#Trans., \#Audio, and \#Emo. denote the numbers of unique transcripts, audio files, and emotion categories, respectively.
\#Motion is the number of unique motion sequences.
\#Motion/Utter. reports the average number of candidate motions per speaker utterance.
Label counts report the numbers of gold/silver/negative candidates (\#$\mathcal{G}$/\#$\mathcal{S}$/\#$\mathcal{N}$).
}
\label{tab:dataset_stats}
\setlength{\tabcolsep}{8pt}
\centering
\resizebox{\linewidth}{!}{
\begin{tabular}{crrrrrrc}
\toprule
\multirow{2}{*}{\textbf{Split}} & \multirow{2}{*}{\textbf{\#Pairs}} &
\multicolumn{3}{c}{\textbf{Speaker Utterance}} &
\multicolumn{1}{c}{\textbf{Listener Reaction}} &
\multicolumn{1}{c}{\textbf{\#Motion/Utter.}} &
\multicolumn{1}{c}{\textbf{Labels (y)}} \\
\cmidrule{3-5}\cmidrule{6-6}\cmidrule{7-8}
& & \textbf{\#Trans.} & \textbf{\#Audio} & \textbf{\#Emo.} &
\textbf{\#Motion} & \textbf{(avg.)} & \textbf{(\#$\mathcal{G}$/\#$\mathcal{S}$/\#$\mathcal{N}$)} \\
\midrule
Train & 137,879 & 6,631 & 6,631 & 46 & 1,822 & 20.79 & 7,527 / 30,862 / 99,490 \\
Val   & 6,790   & 841   & 841   & 40 & 195   & 8.07  & 903 / 1,682 / 4,205 \\
Test  & 6,659   & 826   & 826   & 39 & 197   & 8.06  & 877 / 1,652 / 4,130 \\
\midrule
All   & 151,328 & 8,298 & 8,298 & 47 & 2,029 & 18.24 & 9,307 / 34,196 / 107,825 \\
\bottomrule
\end{tabular}
}
\end{table}

\subsection{Dataset Statistics}
In total, our dataset contains 151{,}328 labeled (speaker utterance, listener reactive motion) pairs, covering 8{,}298 unique speaker utterances and 2{,}029 unique listener reactive motions.
On average, each speaker's utterance is paired with 18.24 candidate reactive motions, highlighting the \textit{one-to-many} nature of listener reactions.
Overall, 9{,}307, 34{,}196, and 107{,}825 pairs are labeled as Gold, Silver, and Negative, respectively, reflecting \textit{graded appropriateness} of candidate reactions.
We split the dataset by \emph{speaker utterance} with an 8:1:1 ratio for train/val/test, such that speaker utterances are \emph{disjoint} across splits (i.e., no utterance appears in more than one split).
Tab.~\ref{tab:dataset_stats} lists detailed statistics.
Our automated construction pipeline further enables straightforward scaling to larger datasets.

\section{Methodology}
\label{sec:methodology}

\begin{figure}[t]
  \centering
  \includegraphics[width=\linewidth]{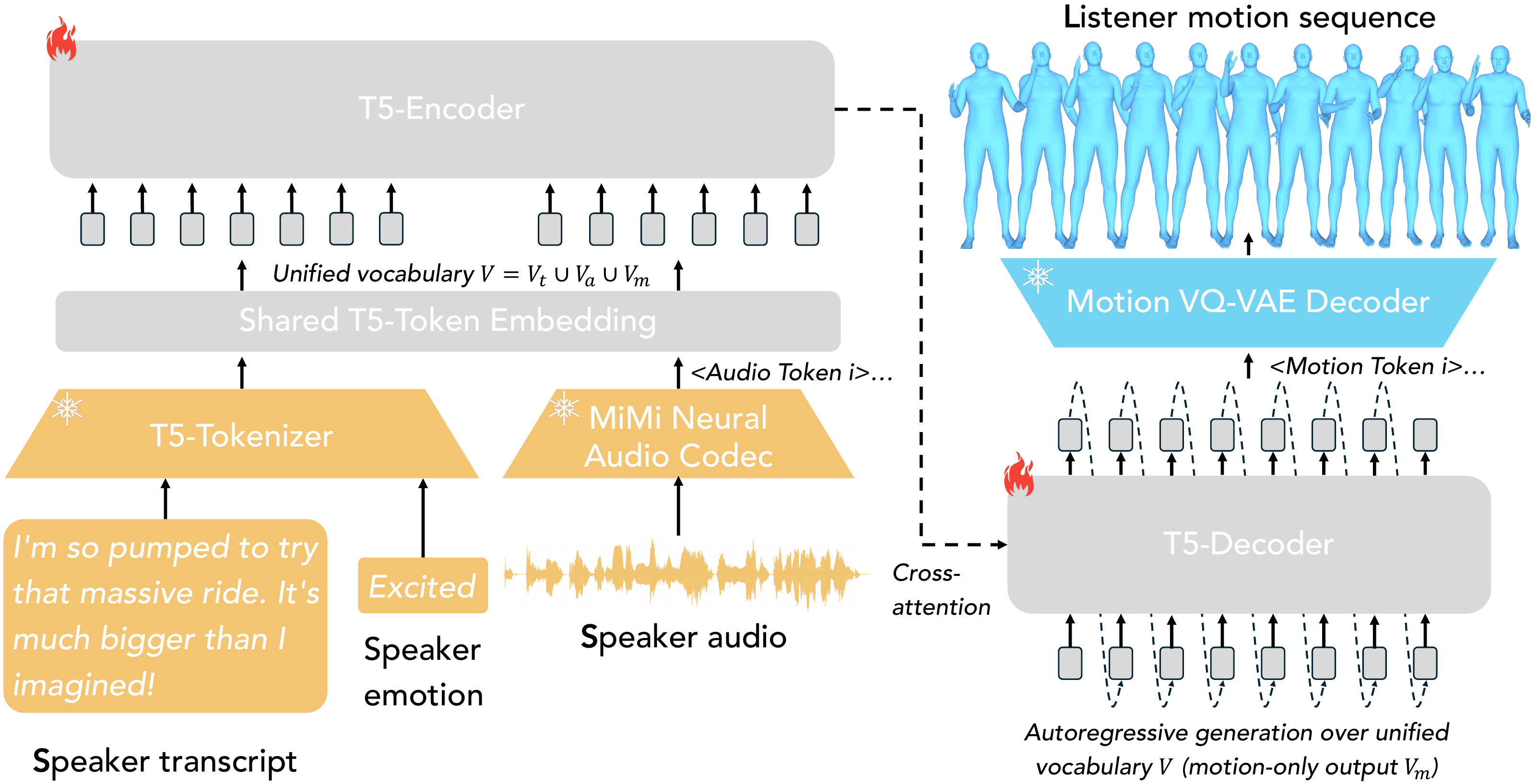}  \setlength{\abovecaptionskip}{-5pt}
  \caption{
    \textbf{Overview of the ReactMotion framework}. 
    We use modality-specific tokenizers to convert raw data, \textit{i.e.}, the speaker’s utterances (including transcript, audio, and emotion) and the listener’s reactive motions, into discrete special tokens. 
    With these tokenizers, a Seq2Seq model is employed to integrate information across modalities and learns to generate the listener’s reactive motions from the speaker’s utterances.
}
\label{fig:overview-arc-reactmotion}
\vspace{-10pt}
\end{figure}

We present ReactMotion, a unified framework for Reactive Listener Motion Generation from Speaker Utterance. 
As illustrated in Fig.~\ref{fig:overview-arc-reactmotion}, we first introduce modality-specific tokenizers that convert raw inputs, \textit{i.e.}, the speaker utterance (including transcript, audio, and emotion) and the listener’s reactive motions, into discrete special tokens. 
With these tokenizers, we employ a Seq2Seq model to unify information across modalities and learn the conditional distribution of the task (Eqn.~\ref{eq:task_def}). 
To capture the one-to-many nature of dyadic interactions, we further train the model with a group-wise preference-based learning objective, which explicitly allows the generation of multiple appropriate reactions for the same speaker utterance.

\subsection{Modality-Specific Tokenization}
\label{sec:tokenization}

We employ modality-specific tokenizers to convert raw data from different modalities into discrete tokens.

\paragraph{Audio Tokenization.}
We use Moshi~\cite{defossez2024moshi} (its Neural
Audio Codec MiMi) to convert the audio waveform in the speaker utterance $A^s$ into discrete codes.
Specifically, its audio encoder $\mathcal{E}_{\text{aud}}(\cdot)$ is employed to extract audio features from $A^s$, which are then quantized using the base codebook $\mathcal{C}_{\text{aud}}$.
\begin{equation}
h_a^s = \mathcal{E}_{\text{aud}}(A^s),
\qquad
x_a^s = \mathcal{Q}_{\text{aud}}(h_a^s),
\label{eq:audio_vq}
\end{equation}
where quantizer $\mathcal{Q}_{\text{aud}}(\cdot)$ maps the features to their nearest entries in the codebook $\mathcal{C}_{\text{aud}}$, and outputs the corresponding codebook indices $x_a^s$.
The resulting indices are treated as discrete audio tokens, allowing the unified model to incorporate audio information while retaining prosody and paralinguistic cues that are informative for reactive behaviors.

\paragraph{Motion Tokenization.}
We represent the listener’s reactive motion $R^l$ as discrete tokens with~\cite{zhang2023generating}, similar to the audio tokenization process:
\begin{equation}
h_m^l = \mathcal{E}_{\text{mot}}(R^l),
\qquad
x_m^l = \mathcal{Q}_{\text{mot}}(h_m^l).
\label{eq:motion_vq}
\end{equation}
where $\mathcal{E}_{\text{mot}}$ and $\mathcal{Q}_{\text{mot}}$ are the motion encoder and quantizer, respectively, and $x_m^l$ are discrete indices of motion codebook $\mathcal{C}_{\text{mot}}$.

Also, the predicted listener reactive motion in the form of discrete tokens from the unified model can be mapped back to the raw motion data through:
\begin{equation}
h_m^l = \mathcal{Q}^{-1}_{\text{mot}}(x_m^l),
\qquad
R^l = \mathcal{D}_{\text{mot}}(h_m^l),
\label{eq:motion_dvq}
\end{equation}
where $\mathcal{Q}^{-1}_{\text{mot}}(\cdot)$ maps the discrete token indices to the vectors in the codebook, and a VQ-VAE motion decoder~\cite{wu2025mg, zhang2023generating} $\mathcal{D}_{\text{mot}}(\cdot)$ decodes the vectors back to the raw motion data.

\subsection{Unified Seq2Seq Modeling}
\label{sec:unified_model}

With above modality-specific tokenizers, we can now represent information across modalities into a unified space, and thus enable a Seq2Seq model to generate a listener reactive motion conditioned on the speaker utterance.

Specifically, we adopt T5-base~\cite{raffel2020exploring} as the Seq2Seq backbone and extend its original textual vocabulary $V_t$ to include audio and motion vocabulary:
\begin{equation}
V = V_t \cup V_m \cup V_a \cup V_s,
\end{equation}
where $V_m$ are the code indices of the motion codebook $\mathcal{C}_{\text{mot}}$, represented as $\{\tokm{<Motion Token i>}\}_{i=0}^{|V_{\mathcal{C}_{\text{mot}}}|-1}$,
and 
 $V_a$ are the code indices of audio codebook $\mathcal{C}_{\text{aud}}$, represented as $\{\toka{<Audio Token i>}\}_{i=0}^{|V_{\mathcal{C}_{\text{aud}}}|-1}$, respectively.
$V_s$ contains special tokens such as $\tokm{<Motion Tokens>}$, $\tokm{</Motion Tokens>}$, $\toka{<Audio Tokens>}$, $\toka{</Audio Tokens>}$, $\toke{<Emotion>}$ and $\toke{</Emotion>}$, which wrap the motion, audio, and emotion token sequences.

This unified vocabulary allows us to formulate reactive listener motion generation, conditioned on different modalities or their combinations $C^s$, in a general format and achieve them within a single model.
Specifically, we first fit discrete codes of the speaker utterance $C^s$ and the listen reactive motion $R^l$ into fixed prompt templates.
Due to page limit, a coarse example task template of using only speaker audio as the condition is shown; detailed one and templates for other conditions are provided in the Appendix~\ref{sec:prompt}.
\begin{tcolorbox}[breakable, boxrule=0pt, colframe=white, sharp corners, left=1mm, right=1mm, top=1mm, bottom=1mm]
\begin{scriptsize}

    \textbf{Input}: You are modeling a speaker-listener dyadic interaction. Given SPEAKER\_AUDIO: [Audio Tokens Placeholder], return ONLY a sequence of listener reactive motion tokens.
    
    \textbf{Output}: [Motion Tokens Placeholder]
    
\end{scriptsize}
\end{tcolorbox}
Now, the modeling process of generating listener reactive motion can be represented as an auto-regressive one, where each motion token is generated with probability
$p_\theta\!\left(x^\text{out}_t \mid x^{\text{in}}(C^s), x^\text{out}_{<t}\right)$.
Here, $x^{\text{in}}(C^s)$ are the input token sequences of the task template embedding with input speaker utterance $C^s$, and $x^{\text{out}}$ are the output token sequences, \textit{i.e.}, listener reactive motion $x_m^l$.

\subsection{Group-wise Preference Learning}
\label{sec:pref_train}

A single speaker utterance $C^s$ can correspond to multiple plausible listener reactive motions $R^l$.
Directly fine-tuning on such one-to-many pairs may lead the model to collapse to averaged and safe behaviors, \textit{e.g.}, standing still.
To mitigate this issue, we train the model using group-wise preference learning.

For each speaker utterance $C^s$, we randomly sample its corresponding listener motions from each label to construct a group
$\{\mathcal{G}, \mathcal{S}, \mathcal{N}\}$,
where $\mathcal{G}$, $\mathcal{S}$, and $\mathcal{N}$ denote the sets of motions labeled as Gold, Silver, and Negative, respectively.
Each motion $R^l$ in the set is represented as a motion token sequence $x_m^l$.
We compute the predicted score for each motion using the length-normalized conditional log-likelihood ~\cite{wu2016google, murray2018correcting, bishop2006pattern}:
\begin{equation}
\ell(x_m^l \mid C^s)
=
\frac{1}{|x_m^l|}
\sum_{t = 1}^{|x_m^l|}
\log p_\theta\!\left(x_{m,t}^l \mid x^{\text{in}}(C^s), x_{m,<t}^l\right).
\label{eq:ll}
\end{equation}
We then aggregate the predicted scores of motions with the same label using a smooth log-mean-exp operator:
\begin{equation}
\ell_{\mathcal{A}}(C^s)
=
\log\!\left(\frac{1}{|\mathcal{A}(C^s)|}
\sum_{x^l2_m \in \mathcal{A}(C^s)}
\exp\big(\ell(x^l_m \mid C^s)\big)\right),
\quad
\mathcal{A}\in\{\mathcal{G},\mathcal{S},\mathcal{N}\}.
\label{eq:set_agg}
\end{equation}
This yields three predicted scores for $C^s$, namely $\ell_{\mathcal{G}}$, $\ell_{\mathcal{S}}$, and $\ell_{\mathcal{N}}$ corresponding to the Gold, Silver, and Negative sets.

Since Gold motions are preferred over Silver, and Silver over Negative, the model is encouraged to produce
$\ell_{\mathcal{G}} > \ell_{\mathcal{S}} > \ell_{\mathcal{N}}$.
We enforce this ordering with a soft-margin ranking loss:
\begin{equation}
\begin{aligned}
\mathcal{L}_{\text{rank}}
&=
\log\!\big(1+\exp\!\big(m-(\ell_{\mathcal{G}}-\ell_{\mathcal{S}})\big)\big)
\;+\;
\log\!\big(1+\exp\!\big(m-(\ell_{\mathcal{S}}-\ell_{\mathcal{N}})\big)\big)
\\
&\quad+\;
\lambda_{gn}\,
\log\!\big(1+\exp\!\big(m-(\ell_{\mathcal{G}}-\ell_{\mathcal{N}})\big)\big),
\end{aligned}
\label{eq:loss_rank}
\end{equation}
where $m$ specifies the margin between different labels, and $\lambda_{gn}$ controls the strength of the Gold$\succ$Negative constraint.

\paragraph{Training objective with frequency reweighting.}
To mitigate the dominance of frequently occurring motion sequences, we apply inverse-frequency weighting based on motion sequence IDs.
Let $i$ index a group (corresponding to one speaker utterance) and let $r_{ij}$ denote the motion sequence ID of the $j$-th candidate in group $i$.
We compute $\mathrm{freq}(r)$ as the number of times motion ID $r$ appears in the \emph{training set} and assign an item weight
$
\tilde{w}_{ij}=\frac{1}{\sqrt{\mathrm{freq}(r_{ij})}}.
\label{eq:item_w}
$
We then define the group weight as the mean item weight within the group,
$w_i=\frac{1}{|\mathcal{C}_i|}\sum_j \tilde{w}_{ij}$, where $\mathcal{C}_i$ denotes the candidate set of group $i$.
Finally, we maximize the aggregated Gold score while applying the ranking loss:
\begin{equation}
\mathcal{L}
=
\frac{\sum_i w_i\Big(-\ell_{\mathcal{G}}^{(i)} + \lambda_{\text{rank}}\mathcal{L}_{\text{rank}}^{(i)}\Big)}
{\sum_i w_i}.
\label{eq:loss_weighted}
\end{equation}

\section{Experiments}
\label{sec:experiments}

\subsection{Implementation Details}
We train  ReactMotion for $100{,}000$ iterations using the default AdamW optimizer and a cosine learning-rate schedule. The learning rate is set to $2\times10^{-5}$ with $1{,}000$ warmup steps. 
We use a per-device batch size of $8$ with gradient accumulation of $2$ steps on a single NVIDIA A100 GPU. 
We train with six conditioning variants ($T$, $A$, $T{+}A$, $T{+}E$,  $A{+}E$, $T{+}A{+}E$) and apply modality dropout ($p{=}0.3$) to improve robustness (see the Appendix~\ref{sec:ex_set_supp} for more details of the implementation).

\subsection{Evaluation Protocol}
\textbf{Evaluation metrics.}  
(i) \textit{Reactive appropriateness}, \ie, how well the generated reactive human motions respond to the speaker’s input, is a core objective of our task. 
Inspired by preference-based evaluation paradigms~\cite{chiang2024chatbot, zheng2023judging, christiano2017deep, bradley1952rank, dubois2024length, stiennon2020learning}, we evaluate reactive appropriateness using group-level win rates \textbf{Win(g$>$G)}, \textbf{Win(g$>$S)}, and \textbf{Win(g$>$N)}. Specifically, we compare the best generated sample $g$ with annotated listener motions labeled as Gold (G), Silver (S), and Negative (N), and compute the win rate against each reference tier.
A win against a higher reference tier (\emph{e.g.,} Silver) indicates that the generated motion is ranked above a higher-quality annotated response, reflecting stronger reactive appropriateness. To realize this evaluation,  we train a \textbf{multimodal judge network} to rank generated reactive body motions conditioned on the same speaker input. Details of the judge network are provided in the appendix.
 We also report \textbf{Gen@3}, the fraction of groups where a generated candidate is ranked within the top-3 among $\{\mathcal{G},\mathcal{S},\mathcal{N}\}$ plus generated candidates under the same group.  
(ii) \textit{Motion quality} is measured by Fr\'echet Inception Distance (FID)~\cite{FID}  computed in a motion feature space, and (iii) \textit{{Diversity}} is measured as the average pairwise embedding distance across generated samples,  following human motion generation \cite{wu2025mg,zhang2023generating}.
(see the Appendix~\ref{sec:supp_metrics} for more details of the evaluation metrics).

\noindent\textbf{Validation of the multimodal judge network.}
Since the judge network is central to measuring reactive appropriateness, we validate it on samples with tiered appropriateness annotations (G/S/N).
Specifically, we compute the tier-consistency win rates \textbf{Win(G$>$S)}, \textbf{Win(G$>$N)}, and \textbf{Win(S$>$N)} to test whether the judge assigns higher scores to more appropriate reactions.
Higher values indicate a more reliable judge.
We further report \textbf{MRR(G)}, which measures how highly the Gold reaction is ranked, and \textbf{nDCG@3}/\textbf{nDCG@5}/\textbf{nDCG@10} to assess graded ranking quality among the top-$K$ candidates.

\begin{table}[t]
\centering
\setlength{\belowcaptionskip}{5pt}
\caption{
\textbf{Multi-modal judge network reliability} under strict modality missingness (Strict-L2).
We evaluate six input modes (text $T$, audio $A$, emotion $E$, and their fusions) on the \textbf{test set}, reporting pairwise win rates (Win(G{>}N), Win(G{>}S), Win(S{>}N)) and ranking metrics (MRR(G), nDCG@K) with graded relevance G>S>N.
}
\label{tab:judge_metrics_strictL2}
\scriptsize
\setlength{\tabcolsep}{3pt}
\renewcommand{\arraystretch}{1.05}
\resizebox{\linewidth}{!}{%
\begin{tabular}{l c c c c c c c}
\toprule
\textbf{Mode} &
\textbf{Win(G$>$N)} $\uparrow$ &
\textbf{Win(G$>$S)} $\uparrow$ &
\textbf{Win(S$>$N)} $\uparrow$ &
\textbf{MRR(G)} $\uparrow$ &
\textbf{nDCG@3} $\uparrow$ &
\textbf{nDCG@5} $\uparrow$ &
\textbf{nDCG@10} $\uparrow$ \\
\midrule
T       & 0.992 & 0.873 & \textbf{0.983} & 0.829 & 0.864 & \textbf{0.878} & 0.932 \\
A       & 0.992 & 0.872 & \textbf{0.983} & \textbf{0.832} & \textbf{0.866} & \textbf{0.878} & \textbf{0.933} \\
T+E     & \textbf{0.993} & 0.876 & 0.982 & 0.826 & 0.857 & 0.876 & 0.929 \\
A+E     & 0.992 & 0.874 & \textbf{0.983} & 0.831 & 0.865 & \textbf{0.878} & \textbf{0.933} \\
T+A     & \textbf{0.993} & \textbf{0.879} & 0.982 & 0.820 & 0.855 & 0.875 & 0.928 \\
T+A+E   & \textbf{0.993} & 0.878 & 0.982 & 0.828 & 0.859 & \textbf{0.878} & 0.930 \\
\bottomrule
\end{tabular}
}
\end{table}

Table~\ref{tab:judge_metrics_strictL2} shows the judge consistently preserves the expected preference ordering with \textbf{near-perfect separation}, across all six modes and both Test set. Gold almost always beats negatives (Win(G$>$N) $\approx$ 0.99) and silver also strongly beats negatives (Win(S$>$N) $\approx$ 0.98), indicating that the judge reliably distinguishes poor motions from plausible ones. Meanwhile, gold beats silver with a clear margin (Win(G$>$S) $\approx$ 0.87--0.88), reflecting sensitivity to fine-grained quality differences beyond simply rejecting negatives. The judge further achieves strong ranking quality (MRR(G) $\approx$ 0.82--0.84; nDCG@5 $\approx$ 0.87--0.88; nDCG@10 $\approx$ 0.93), demonstrating stable and meaningful top-$K$ ordering.

Although our multimodal judge network is trained on multiple input modalities, \ie, text ($T$), audio ($A$), and emotion ($E$), it supports missing modalities using Strict-L2. Disabled modalities are replaced with information-free inputs (all-padding text, all-padding audio codes, or an unknown emotion token).
This enables the judge network to operate with any subset of modalities; even with a single modality, it performs well in evaluation.
(see the Appendix~\ref{sec:judge_network} and ~\ref{sec:judge_impl} for more details of the judge network).

\subsection{Quantitative Results}

Since reactive listener motion generation remains underexplored, we evaluate a set of representative baselines.
\textbf{(a) Random Selection} uniformly samples a motion sequence from \textsc{HumanML3D}~\cite{guo2022generating}.
\textbf{(b) Retrieval} applies the text--motion matching network from prior \textsc{HumanML3D} T2M work~\cite{wu2025mg, zhang2023generating} to compute text--motion similarity and retrieves the nearest-neighbor listener motion sequence from the training set given the speaker transcript.
We also consider stronger \textbf{cascaded LLM$\rightarrow$T2M} baselines: given a speaker utterance (and emotion), an LLM~\cite{qwen3} first generates a listener-motion caption, which is then passed to a T2M generator to synthesize the final motion.
We instantiate the LLM with \textsc{Qwen3-30B-A3B} (30.5B parameters) and a fine-tuned \textsc{Qwen3-4B-Thinking} (4B parameters) trained on our training-set (speaker utterance, listener-motion caption) pairs.
The resulting captions are fed into two representative T2M generators, \textsc{T2M-GPT}~\cite{zhang2023generating} and \textsc{MG-MotionLLM}~\cite{wu2025mg}.
More details of baselines are in the Appendix~\ref{sec:baseline_network}.

Tab.~\ref{tab:main_results} shows that ReactMotion outperforms all baselines in reactive appropriateness.
Among the cascaded LLM$\rightarrow$T2M pipelines, LLM$\rightarrow$\textsc{MG-MotionLLM~\textsuperscript{*}} is the strongest, improving over Random Selection and Retrieval.
However, despite using a powerful motion generator, it still performs poorly under strict comparisons to Silver references (Win(g$>$S)), indicating that the two-stage caption-then-generate pipeline struggles to produce highly appropriate listener reactions.

In contrast, ReactMotion achieves near-perfect Win(g$>$N) across input modes and substantially improves Win(g$>$S) and Gen@3.
Our full model ($T{+}A{+}E$) yields the best overall Win rates, while maintaining low FID and competitive diversity.
Although Retrieval attains the highest diversity by construction, it yields much lower appropriateness and worse realism than our approach.
More experimental results are provided in the Appendix~\ref{sec:experiment}.

\begin{table}[t]
\setlength{\belowcaptionskip}{5pt} 
\caption{\textbf{Quantitative results} on the test set. Main evaluation metrics are Win(g$>$N), Win(g$>$S), Win(g$>$G), Gen@3  measuring Reactive Appropriateness. We additionally evaluate motion quality (FID) and diversity. $^{*}$ indicates that the LLM is fine-tuned using training-set speaker utterance and listener motion caption pairs.}  
\label{tab:main_results}
\centering
\resizebox{\linewidth}{!}{
\begin{tabular}{l l c c c  c c c}
\toprule
Method & Input Mod.  &
{Win(g$>$N)}$\uparrow$ &
{Win(g$>$S)}$\uparrow$ &
{Win(g$>$G)}$\uparrow$ &
Gen@3$\uparrow$ &
FID$\downarrow$ &
Diversity$\uparrow$ \\
\midrule
GT & - & -   &-   &-   &-    &0.278  &6.187 \\
Random Selection & -  &0.265 &0.122 &0.006 &0.099  &42.363 &9.880 \\

\midrule
Retrieval & $T$  & 0.392 &0.252 &0.130 &0.206 &7.429 &\textbf{8.207} \\

\shortstack[l]{LLM$\rightarrow$\textsc{T2M-GPT}} 
& $T{+}E$  & 0.138 & 0.038 & 0.016 & 0.199 & 49.920 & 4.946 \\
\shortstack[l]{LLM$\rightarrow$\textsc{T2M-GPT}}~\textsuperscript{*} 
& $T{+}E$& 0.171 & 0.027 & 0.017 & 0.350 & 42.589 & 6.102 \\

\shortstack[l]{LLM$\rightarrow$\textsc{MG-MotionLLM}}& $T{+}E$ & 0.775 & 0.245 & 0.044 & 0.345 & 23.629 & 5.082 \\
\shortstack[l]{LLM$\rightarrow$\textsc{MG-MotionLLM}}~\textsuperscript{*}
& $T{+}E$  & 0.883 & 0.274 & 0.047 & 0.380 & 25.723 & 4.546 \\

\midrule
ReactMotion (Ours) & $T$  & 0.993 &0.774 & 0.258  & 0.916 & \textbf{4.706} &  4.789\\
ReactMotion (Ours) & $A$  & 0.992 & 0.614 & 0.164 & 0.864 & 6.221 & 4.009 \\
ReactMotion (Ours) & $T{+}E$ & 0.990 & 0.696 & 0.206 & 0.930  & 5.422 & 4.475 \\
ReactMotion (Ours) & $A{+}E$  & 0.993 & 0.736 & \textbf{0.323} & \textbf{0.981}  & 6.485 & 4.162 \\
ReactMotion (Ours)& $T{+}A$& 0.993 & 0.651 & 0.215 & 0.931 & 6.560 & 4.145 \\
ReactMotion (Ours) & $T{+}A{+}E$  & \textbf{1.000} & \textbf{0.797} & 0.266 & 0.960 & 4.760 & 4.804 \\
\bottomrule
\end{tabular}}
\end{table}

\begin{figure}[t]
  \centering
  \includegraphics[width=0.9\linewidth]{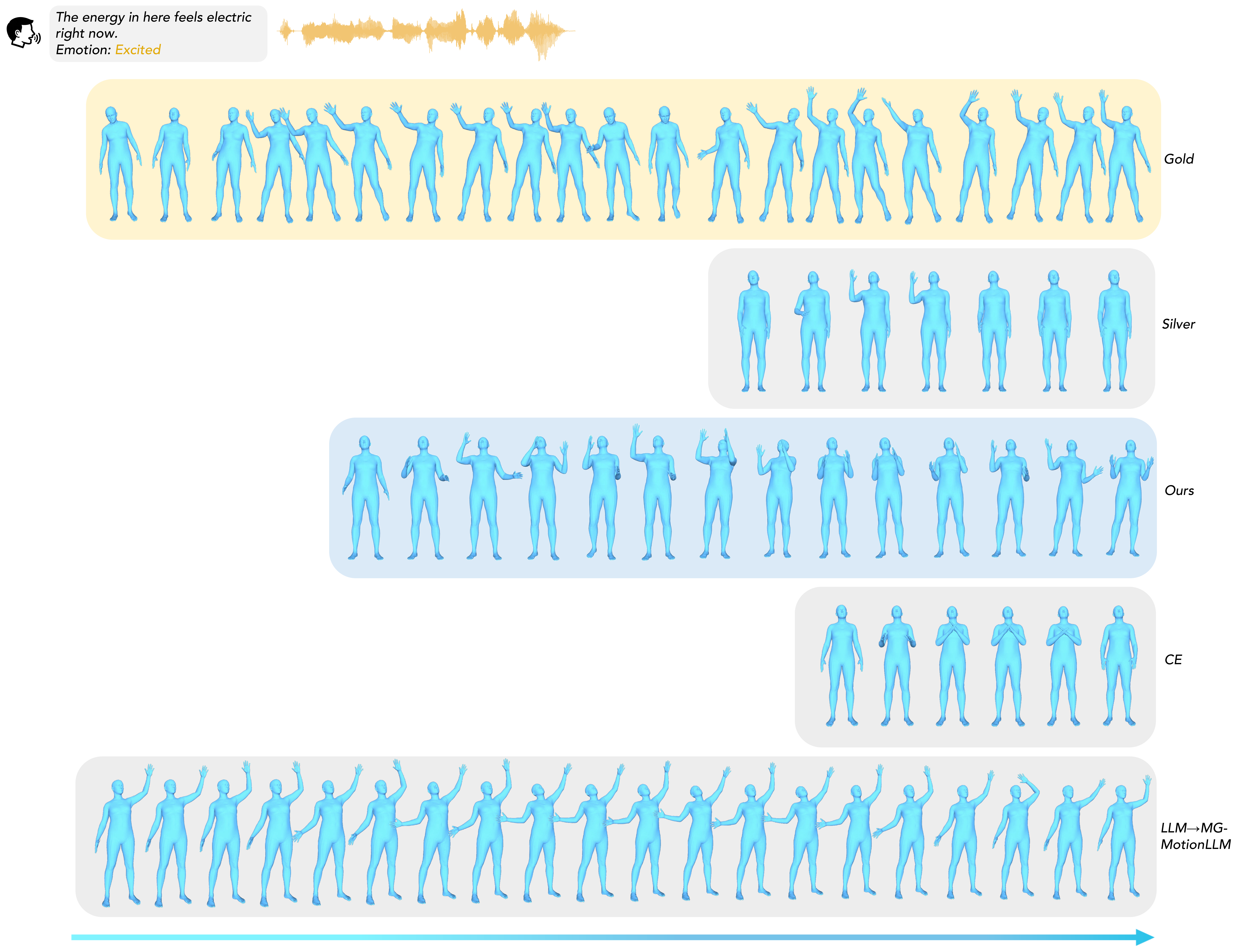}
  \caption{\textbf{Qualitative results.} We compare gold and silver listener reactions, motions generated by our ReactMotion (Ours), a cross-entropy trained variant (CE), and a cascaded LLM$\rightarrow$T2M baseline, all conditioned on the same speaker utterance. We visualize the resulting 3D motion sequences.}
  \label{fig:vis}
\end{figure}

\subsection{Qualitative Results}

We visualize representative examples in Fig.~\ref{fig:vis}, comparing our ReactMotion (Ours), a cross-entropy trained variant (CE), and LLM$\rightarrow$MG-MotionLLM~\textsuperscript{*} with a finetuned Qwen~\cite{qwen3} (\textsc{Qwen3-4B-Thinking}) on training set, together with gold and silver reference reactions under the same speaker condition.
Overall, \textbf{ReactMotion} produces reactive motions that are both semantically consistent with the speaker content and expressive in intensity.
For instance, for the utterance ``The energy in here feels electric right now'' with \emph{excited} emotion, our model generates larger, more dynamic upper-body and arm movements, which better reflect the high-energy ``electric'' cue and match the communicative style seen in the gold reaction.

In contrast, the \textbf{silver} reaction exhibits a rapid hand-wave but remains relatively low-energy, making it less aligned with the excited condition.
The \textbf{CE} variant tends to regress to generic, weakly-conditioned responses (\emph{e.g,} a static pose such as crossing arms), indicating limited ability to exploit preference structure and model the one-to-many nature of reactive behaviors.
Finally, the \textbf{LLM$\rightarrow$T2M} baseline often generates repetitive motions (\emph{e.g,} near-constant waving) with limited temporal variation, which appears less suitable for dyadic communication, where reactions typically evolve over time (\emph{e.g,} hands rising and lowering, pose changes and subtle turns).
Moreover, because dyadic reactions can be difficult to describe in natural language, the out-of-domain captions produced by the LLM may be noisy, which can lead MG-MotionLLM to produce degraded outputs, including overly short motion sequences.

\subsection{User Study}
\begin{figure}[t]
  \centering
  \includegraphics[width=0.7\linewidth]{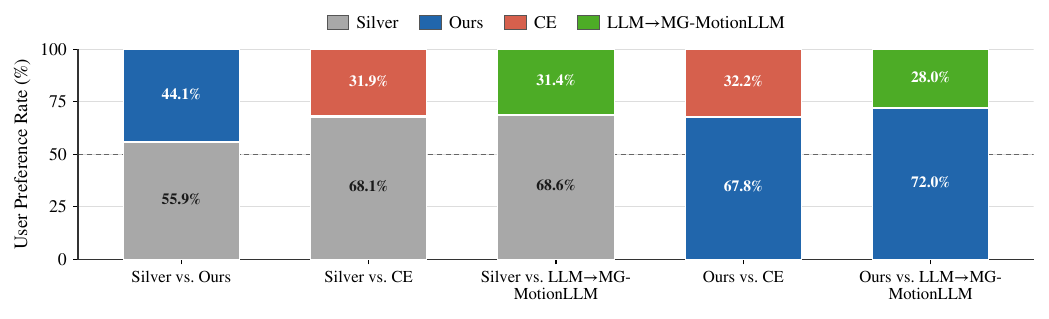}
  \setlength{\abovecaptionskip}{5pt} 
  \caption{\textbf{User study} on reactive appropriateness. }
  \label{fig:user_study}
\end{figure}

We recruit 59 volunteers and conduct a user study to evaluate the reactive appropriateness of listener motions generated by \textbf{ReactMotion} (\textbf{Ours}) against two baselines (the \textbf{CE} variant and \textbf{LLM}$\rightarrow$\textbf{MG-MotionLLM}~\textsuperscript{*}) and the \emph{best-in-group} \textbf{Silver} reference.
In each case, participants watch two motion videos (A/B) conditioned on the same speaker utterance (audio with transcript shown) and select the more appropriate listener reaction.
Each participant completes 36 cases covering six speaker conditions (six pairwise comparisons per condition).

As shown in Fig.~\ref{fig:user_study}, \textbf{Ours} is preferred over the generative baselines, achieving win rates of 67.8\% against \textbf{CE} and 72.0\% against \textbf{LLM}$\rightarrow$\textbf{MG-MotionLLM}~\textsuperscript{*}.
\textbf{Ours} is also competitive with the \textbf{Silver} reference, receiving 44.1\% of votes in \textbf{Silver} vs.\ \textbf{Ours}, substantially higher than \textbf{CE} (31.9\%) and \textbf{LLM}$\rightarrow$\textbf{MG-MotionLLM} (31.4\%).

\begin{table}[t]
\centering
\setlength{\belowcaptionskip}{5pt}
\caption{
\textbf{Ablation studies} on the test split (all use $A{+}T{+}E$ unless noted).
\textit{w/o} denotes training without the corresponding component.
The \textbf{CE baseline} trains the same model using only a cross-entropy loss by pairing each speaker input with a single Gold reaction as supervision.
}
\label{tab:ablation}
\resizebox{\linewidth}{!}{
\begin{tabular}{l c c c c c c}
\toprule
Method  &
{Win(g$>$N)}$\uparrow$ &
{Win(g$>$S)}$\uparrow$ &
{Win(g$>$G)}$\uparrow$ &
Gen@3$\uparrow$ &
FID$\downarrow$ &
Diversity$\uparrow$ \\
\midrule
CE baseline & 0.990 & 0.741 & 0.262 & 0.938 & 6.555 & 5.448 \\
\textbf{Ours (full)}   & \textbf{1.000} & \textbf{0.797}  &  \textbf{0.266} & \textbf{0.960} & \textbf{4.760} & 4.804 \\
\textit{w/o} Inverse-frequency reweighting & 0.979 & 0.704 & 0.220 & 0.946 & 5.177 & 4.929 \\
\textit{w/o} $\mathcal{L}_{\text{rank}}$   &0.996 & 0.781 & 0.260 & \textbf{0.960} & 5.950 & \textbf{5.453} \\
\textit{w/o} $\ell_{\mathcal{G}}$ & 0.996 & 0.712 & 0.215 & 0.943 & 6.376 & 4.493 \\
\bottomrule
\end{tabular}}
\end{table}

\subsection{Ablation Studies}
\label{sec:ablation_studies}

\paragraph{Modality study.} 
We study the effect of input modalities in Tab.~\ref{tab:main_results}.
Across settings, multimodal fusion performs best overall.
Text is the strongest single cue, giving high alignment and the lowest single-modality FID (\emph{e.g.,} $T$: Win(g{>}N)=0.993, Win(g{>}S)=0.774, FID=4.706).
Audio alone is weaker for fine-grained appropriateness, but adding emotion substantially improves it (best Win(g{>}G)=0.323 and Gen@3=0.981).
Full fusion ($T{+}A{+}E$) is the most balanced, achieving the best Win(g{>}N)=1.000, strong Win(g{>}S)=0.797, and a low FID=4.760.

\paragraph{Ablations on group-wise preference learning.}
Tab.~\ref{tab:ablation} ablates key components of our group-wise preference learning objective.
Compared to training with cross-entropy only, our \textbf{full model} substantially improves both reactive appropriateness and motion quality (e.g., Win(g$>$S): 0.741$\rightarrow$0.797; Gen@3: 0.938$\rightarrow$0.960; FID: 6.555$\rightarrow$4.760).
Removing \textbf{inverse-frequency reweighting} leads to the largest appropriateness drop, especially against the strongest tier (Win(g$>$G): 0.266$\rightarrow$0.220), highlighting the importance of mitigating the dominance of frequent and generic motions.
Removing the \textbf{ranking loss} degrades fidelity (FID: 4.760$\rightarrow$5.950) while increasing diversity (4.804$\rightarrow$5.453), suggesting that the ranking constraints help enforce correct relative ordering among tiers.
Finally, removing $\ell_{\mathcal{G}}$ consistently harms both appropriateness and quality, indicating that likelihood supervision on Gold reactions remains necessary.

\section{Conclusion}
\label{sec:conclusion}
We introduce \emph{Reactive Listener Motion Generation from Speaker Utterance}, a new task for modeling listener motion responses in dyadic interactions. To support this task, we present \textbf{ReactMotionNet}, a multi-modal dataset that explicitly captures the inherent non-determinism of human behavior: for each speaker utterance, we provide multiple candidate listener motions with preference annotations, enabling supervision beyond a single ``ground-truth'' response. Building on this dataset design, we develop preference-oriented evaluation protocols tailored to reactive motion generation. Finally, we propose \textbf{ReactMotion}, a unified framework that processes multi-modal speaker cues, substantially outperforms strong baselines in motion quality and reactive appropriateness. We believe this work provides a foundation for future research on modeling dyadic interactions.

\clearpage
\setcounter{section}{0}
\renewcommand{\thesection}{\Alph{section}}
\section*{Outline of the Supplementary Material}

The supplementary material is organized as follows:

\begin{itemize}[label=\textbullet, leftmargin=1.5em]
    \item \textbf{Section~\ref{sec:ex_set_supp}} presents the implementation details, including the model configuration, vocabulary construction, optimization settings, and training hyperparameters.
    \item \textbf{Section~\ref{sec:model_efficiency}} presents the model size of ReactMotion.
     \item \textbf{Section~\ref{sec:prompt}}: prompt templates for different speaker-condition settings;
    \item \textbf{Section~\ref{sec:methodological_details}} further provides the additional evaluation details, including:
    \begin{itemize}[label=\textbullet, leftmargin=1.5em]
       
        \item \textbf{Section~\ref{sec:judge_network}}: the formulation of the multimodal judge network;
        \item \textbf{Section~\ref{sec:baseline_network}}: details of the baseline methods.
    \end{itemize}

    \item \textbf{Section~\ref{sec:supp_metrics}} introduces the evaluation metrics, covering reactive appropriateness, motion quality, and diversity.

    \item \textbf{Section~\ref{sec:supp_dataset}} provides additional statistics and analysis of the ReactMotionNet dataset.

    \item \textbf{Section~\ref{sec:supp_hparam}} presents the hyperparameter sensitivity analysis, including the full sweep results, representative configurations, and heatmap visualizations. 
    
   \item  \textbf{Section} \textbf{\ref{Sec:inference}}   evaluates the inference efficiency of the proposed method. 
    \item \textbf{Section~\ref{sec:supp_user_study}} reports the protocol and results of the user study.

    \item \textbf{Section~\ref{sec:supp_failure}} shows representative failure cases.

    \item \textbf{Section~\ref{sec:supp_limitations}} discusses the limitations of the current framework.
\end{itemize}

\section{Implementation Details}
\label{sec:ex_set_supp}

{\setlength{\tabcolsep}{14pt}
\begin{table}[h]
\caption{Implementation details and hyperparameters used in training.}
\centering
\resizebox{0.9\textwidth}{!}{
\begin{tabular}{ll}
\toprule
\textbf{Setup} & \textbf{Value} \\
\midrule
Seq2Seq backbone model & T5-base~\cite{raffel2020exploring} \\
Text tokenizer & T5-base tokenizer~\cite{raffel2020exploring} \\
Audio tokenizer & MiMi neural audio codec~\cite{defossez2024moshi} \\
Motion tokenizer & VQ-VAE from T2M-GPT~\cite{zhang2023generating} \\
\midrule
Per-device batch size & 8 \\
Gradient accumulation steps & 2 \\
Training steps & 100,000 \\
Warmup steps & 1,000 \\
Optimizer & AdamW \\
Adam $\beta_1$ & 0.9 \\
Adam $\beta_2$ & 0.999 \\
Weight decay & 0.0 \\
Learning rate & $2.0 \times 10^{-5}$ \\
\midrule
Maximum  source length & 512 \\
Maximum  target length & 256 \\
Text vocabulary size $|V_t|$ & 32,100 \\
Audio codebook size $|V_a|$ & 2,048 \\
Number of MiMi audio codebooks & 8 \\
Motion VQ-VAE codebook size $|V_m|$ & 512 \\
Total vocabulary size $|V|$ & 49,002 \\
\midrule
Backbone parameters & 222.9M \\
Total trainable parameters after vocabulary expansion & 235.9M \\
\midrule
Ranking loss weight $\lambda_{\text{rank}}$ & 0.25 \\
Gold-negative loss weight $\lambda_{\text{gn}}$ & 0.25 \\
Ranking Margin $m$ & 0.5 \\
Modality dropout rate & 0.3 \\
LogSumExp normalization & Enabled \\ 
\bottomrule
\end{tabular}
}
\label{tb:hyperparam_details}
\end{table}
}

Tab.~~\ref{tb:hyperparam_details} summarizes the key implementation details and training hyperparameters used in our experiments.
Specifically, ReactMotion is instantiated with a T5-base Seq2Seq backbone, comprising 222.9M backbone parameters and 235.9M trainable parameters after extending the vocabulary.
In accordance with the methodology section, the original textual vocabulary ($|V_t|=32{,}100$) is augmented with motion tokens ($|V_m|=512$), MiMi audio tokens ($|V_a|=2{,}048$ per codebook; 8 codebooks), and modality-specific special tokens that mark the boundaries of different modalities, resulting in a unified vocabulary of size $63{,}338$. Notably, the vocabulary includes tokens from all 8 MiMi codebooks for completeness, while in practice we only use tokens from the base codebook during training to accelerate the process.
The model takes tokenized speaker utterances as input and autoregressively predicts listener reactive motion tokens, with maximum source and target lengths set to 512 and 256, respectively.
We train the model using AdamW with learning rate $2.0\times10^{-5}$, $\beta_1=0.9$, $\beta_2=0.999$, weight decay 0.0, 1{,}000 warmup steps, per-device batch size 8, gradient accumulation over 2 steps, and 100{,}000 total optimization steps.
To capture the one-to-many mapping from a speaker utterance to plausible listener reactions, training adopts the proposed group-wise preference objective with $\lambda_{\text{rank}}=0.25$, $\lambda_{\text{gn}}=0.25$, and margin $m=0.5$.
We further apply modality dropout with rate 0.3 to improve robustness to missing modalities, while length-normalized LogSumExp aggregation is used to obtain stable set-level scores during preference optimization.

\subsection{Model Size}
\label{sec:model_efficiency}

\begin{table}[h]
\centering
\caption{Model Configuration and Parameters of ReactMotion.}
\resizebox{0.4\linewidth}{!}{
\begin{tabular}{lc}
\toprule
\textbf{Metric} & \textbf{Value} \\
\midrule
Backbone parameters & 222.9M \\
Total trainable parameters & 235.9M \\
Unified vocabulary size & 49,002 \\
\bottomrule
\end{tabular}
}
\label{tab:size}
\end{table}

Table~\ref{tab:size} summarizes the model size of ReactMotion.
The model is built upon a T5-base backbone with 222.9M parameters and 235.9M trainable parameters after extending the vocabulary to incorporate multimodal tokens.

\subsection{Prompt Templates}
\label{sec:prompt}

To support unified generation under different speaker-condition settings, we convert the available speaker cues into a fixed natural-language prompt template.
Given a speaker utterance consisting of transcription, audio, and optional emotion annotation, we construct the input prompt by selectively enabling the corresponding fields.
The model is instructed to output \emph{only} the listener motion-token sequence in a strict format, without any additional natural language.

Formally, for a speaker utterance $C^s$, the prompt is constructed as
\begin{tcolorbox}[breakable, boxrule=0pt, colframe=white, sharp corners, left=1mm, right=1mm, top=1mm, bottom=1mm]
\begin{scriptsize}
\textbf{Input:}

You are modeling a speaker-listener dyadic interaction.

Input:

- SPEAKER\_TRANSCRIPTION: [Speaker Transcription]

- SPEAKER\_AUDIO: [Speaker Audio]

- SPEAKER\_EMOTION: <Emotion> [Speaker Emotion] </Emotion>

Output:

Return ONLY a sequence of listener motion tokens in the exact format:

<Motion Tokens> <Motion Token i> ... </Motion Tokens>

Do NOT output any other words.
\end{scriptsize}
\end{tcolorbox}

In practice, the fields in the prompt are enabled or disabled depending on the chosen condition mode.
For example, when transcription is used but audio is not, the \texttt{SPEAKER\_AUDIO} field is left empty; when emotion is disabled, the emotion line is omitted entirely.
This design allows us to handle text-only, audio-only, text+audio, text+emotion, audio+emotion, and text+audio+emotion settings within a single unified framework.

Below we show several concrete examples.

\paragraph{Text-only condition ($T$).}
\begin{tcolorbox}[breakable, boxrule=0pt, colframe=white, sharp corners, left=1mm, right=1mm, top=1mm, bottom=1mm]
\begin{scriptsize}
\textbf{Input:}

You are modeling a speaker-listener dyadic interaction.

Input:

- SPEAKER\_TRANSCRIPTION: [Speaker Transcription]

- SPEAKER\_AUDIO: 

Output:

Return ONLY a sequence of listener motion tokens in the exact format:

<Motion Tokens> <Motion Token i> ... </Motion Tokens>

Do NOT output any other words.
\end{scriptsize}
\end{tcolorbox}

\paragraph{Text+Emotion condition ($T{+}E$).}
\begin{tcolorbox}[breakable, boxrule=0pt, colframe=white, sharp corners, left=1mm, right=1mm, top=1mm, bottom=1mm]
\begin{scriptsize}
\textbf{Input:}

You are modeling a speaker-listener dyadic interaction.

Input:

- SPEAKER\_TRANSCRIPTION: [Speaker Transcription]

- SPEAKER\_AUDIO: 

- SPEAKER\_EMOTION: <Emotion> [Speaker Emotion] </Emotion>

Output:

Return ONLY a sequence of listener motion tokens in the exact format:

<Motion Tokens> <Motion Token i> ... </Motion Tokens>

Do NOT output any other words.
\end{scriptsize}
\end{tcolorbox}

\paragraph{Audio-only condition ($A$).}
\begin{tcolorbox}[breakable, boxrule=0pt, colframe=white, sharp corners, left=1mm, right=1mm, top=1mm, bottom=1mm]
\begin{scriptsize}
\textbf{Input:}

You are modeling a speaker-listener dyadic interaction.

Input:

- SPEAKER\_TRANSCRIPTION: 

- SPEAKER\_AUDIO: [Speaker Audio]

Output:

Return ONLY a sequence of listener motion tokens in the exact format:

<Motion Tokens> <Motion Token i> ... </Motion Tokens>

Do NOT output any other words.
\end{scriptsize}
\end{tcolorbox}

\paragraph{Audio+Emotion condition ($A{+}E$).}
\begin{tcolorbox}[breakable, boxrule=0pt, colframe=white, sharp corners, left=1mm, right=1mm, top=1mm, bottom=1mm]
\begin{scriptsize}
\textbf{Input:}

You are modeling a speaker-listener dyadic interaction.

Input:

- SPEAKER\_TRANSCRIPTION: 

- SPEAKER\_AUDIO: [Speaker Audio]

- SPEAKER\_EMOTION: <Emotion> [Speaker Emotion] </Emotion>

Output:

Return ONLY a sequence of listener motion tokens in the exact format:

<Motion Tokens> <Motion Token i> ... </Motion Tokens>

Do NOT output any other words.
\end{scriptsize}
\end{tcolorbox}

\paragraph{Text+Audio condition ($T{+}A$).}
\begin{tcolorbox}[breakable, boxrule=0pt, colframe=white, sharp corners, left=1mm, right=1mm, top=1mm, bottom=1mm]
\begin{scriptsize}
\textbf{Input:}

You are modeling a speaker-listener dyadic interaction.

Input:

- SPEAKER\_TRANSCRIPTION: [Speaker Transcription]

- SPEAKER\_AUDIO: [Speaker Audio]

Output:

Return ONLY a sequence of listener motion tokens in the exact format:

<Motion Tokens> <Motion Token i> ... </Motion Tokens>

Do NOT output any other words.
\end{scriptsize}
\end{tcolorbox}

\paragraph{Text+Audio+Emotion condition ($T{+}A{+}E$).}
\begin{tcolorbox}[breakable, boxrule=0pt, colframe=white, sharp corners, left=1mm, right=1mm, top=1mm, bottom=1mm]
\begin{scriptsize}
\textbf{Input:}

You are modeling a speaker-listener dyadic interaction.

Input:

- SPEAKER\_TRANSCRIPTION: [Speaker Transcription]

- SPEAKER\_AUDIO: [Speaker Audio]

- SPEAKER\_EMOTION: <Emotion> [Speaker Emotion] </Emotion>

Output:

Return ONLY a sequence of listener motion tokens in the exact format:

<Motion Tokens> <Motion Token i> ... </Motion Tokens>

Do NOT output any other words.
\end{scriptsize}
\end{tcolorbox}

Given the constructed prompt $x^{\text{in}}(C^s)$, the model auto-regressively predicts the listener motion-token sequence $x^{\text{out}}$ as
\[
p_\theta\!\left(x_t^{\text{out}} \mid x^{\text{in}}(C^s), x_{<t}^{\text{out}}\right).
\]
Here, $x^{\text{in}}(C^s)$ denotes the prompt sequence instantiated from the speaker utterance $C^s$, and $x^{\text{out}}$ denotes the output listener motion-token sequence.

\section{Additional Evaluation Details}
\label{sec:methodological_details}

\subsection{Multimodal Judge Network}
\label{sec:judge_network}

\begin{figure}[h!]
    \centering
    \includegraphics[width=\linewidth]{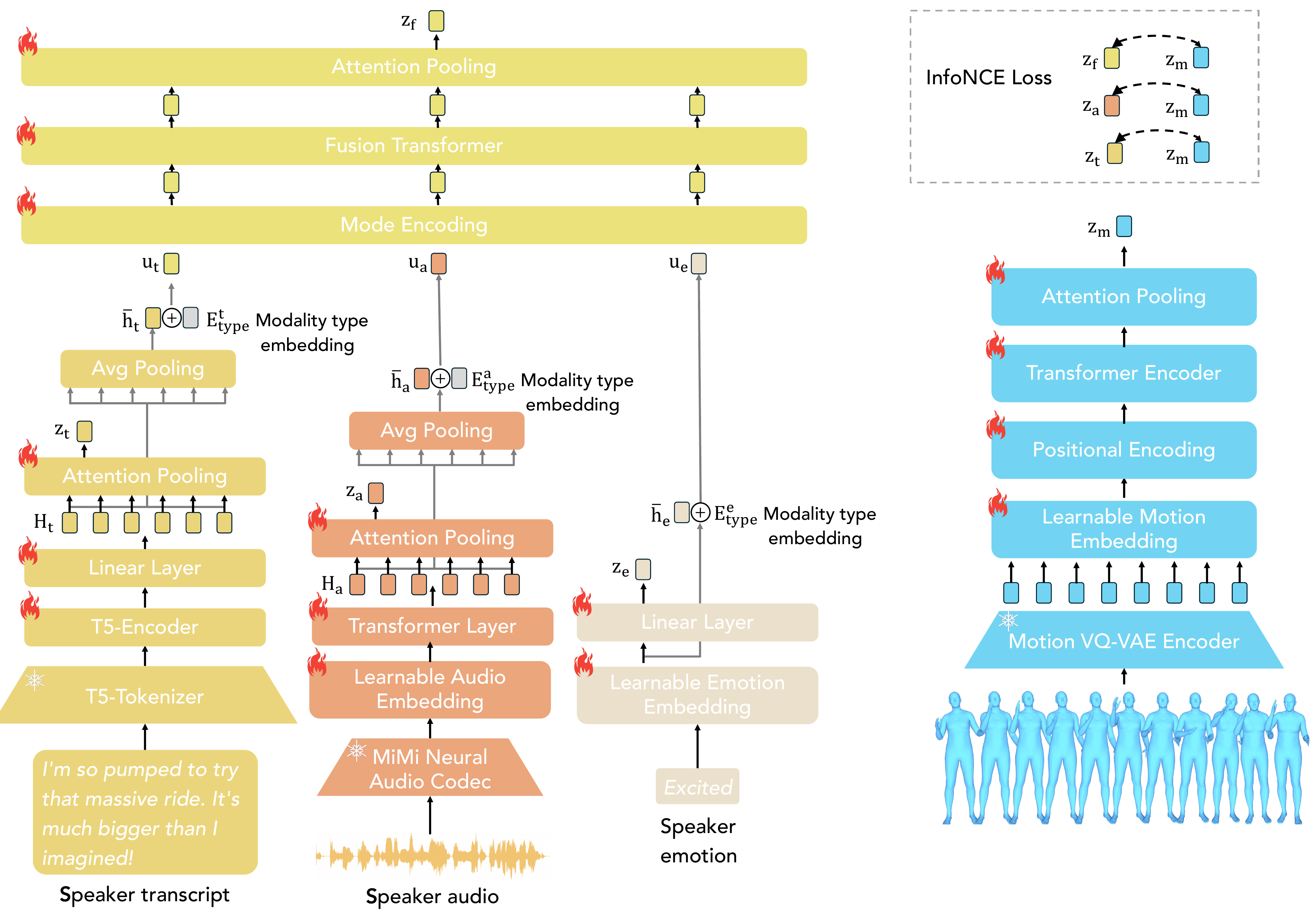}
    \setlength{\abovecaptionskip}{-5pt}
    \caption{
    \textbf{Architecture of the multimodal judge network.}
    Given a speaker utterance and a candidate listener motion, the judge encodes transcript, MiMi audio tokens, and the discrete emotion label with three modality-specific branches, producing modality embeddings \(z_t\), \(z_a\), and \(z_e\), as well as hidden summaries used to form fusion tokens \(u_t\), \(u_a\), and \(u_e\).
    Modality-type embeddings and a mode embedding are added to these tokens, which are then processed by a fusion transformer and attention pooling to obtain the unified condition embedding \(z_f\).
    In parallel, the candidate listener motion, represented by VQ-VAE motion tokens, is encoded by a motion transformer and pooled into a motion embedding \(z_m\).
    The judge computes compatibility between the condition and motion embeddings in a shared normalized scoring space.
    During training, a group-wise InfoNCE objective is applied to the fused embedding and auxiliary modality-specific embeddings, enabling reliable scoring under both full and partial speaker utterances.
    Snowflake and flame icons denote frozen and trainable modules, respectively.
    }
    \label{fig:judge_network}
\end{figure}

To evaluate the reactive appropriateness of generated listener motions and support best-of-\(K\) selection, we train a multimodal judge network, illustrated in Fig.~\ref{fig:judge_network}.
Given a speaker utterance \(C^s\) and a candidate listener motion token sequence \(x_m^l\), the judge network \(s_\psi\) outputs a scalar compatibility score
\begin{equation}
s_\psi(C^s, x_m^l) \in \mathbb{R},
\end{equation}
where a larger value indicates that the candidate listener motion is more appropriate for the given speaker utterance.

\paragraph{Architecture.}
It contains three branches to encode different modalities: transcript, audio, and emotion in the speaker utterance \(C^s\), a fusion branch that integrates the available information in \(C^s\) while allowing missing modalities, and a motion branch to encode the reactive motion.
All branches project their features from dimension \(d\) into a shared scoring space of dimension \(d_o\).
By default, all score-space embeddings are \(\ell_2\)-normalized.

\textbf{Text branch.}
Let \(x_t^s\) denote the tokenized speaker transcript, and let \(M_t \in \{0,1\}^{T_t}\) denote its padding mask, where \(M_t(j)=1\) indicates that the \(j\)-th token is valid and \(M_t(j)=0\) indicates padding.
We encode the transcript with a T5 encoder \(\mathcal{E}_{\mathrm{T5}}(\cdot)\) and project the resulting hidden states into the shared hidden space:
\begin{equation}
H_t = W_t\,\mathcal{E}_{\mathrm{T5}}(x_t^s) + b_t,
\label{eq:judge_text_enc}
\end{equation}
where \(H_t \in \mathbb{R}^{T_t \times d}\), \(W_t \in \mathbb{R}^{d \times d_{\mathrm{T5}}}\), and \(b_t \in \mathbb{R}^{d}\) are learnable parameters.
We then aggregate the token-level features into a text embedding in the scoring space:
\begin{equation}
\tilde{z}_t = \mathrm{AttnPool}_t(H_t; M_t),
\qquad
z_t = \mathrm{L2Norm}(\tilde{z}_t),
\label{eq:judge_text_pool}
\end{equation}
where \(\tilde{z}_t,z_t \in \mathbb{R}^{d_o}\), \(\mathrm{AttnPool}_t(\cdot)\) denotes a masked attention-pooling operator that ignores padded positions according to \(M_t\),
and $ \mathrm{L2Norm}$ denotes $\ell_{2}$ normalization.

\vspace{1em}
\textbf{Audio branch.}
Let \(x_a^s\) denote the speaker audio token sequence obtained from the MiMi neural codec tokenizer~\cite{defossez2024moshi}.
Because MiMi audio is represented by multiple codebooks, we first map the discrete tokens into embeddings, add learnable codebook-level embeddings and positional embeddings, and then process the resulting sequence with a transformer encoder:
\begin{equation}
H_a = \mathcal{E}_a\!\big(\mathrm{Emb}_a(x_a^s) + E_{\mathrm{lvl}}^{a} + E_{\mathrm{pos}}^{a}\big),
\label{eq:judge_audio_enc}
\end{equation}
where \(H_a \in \mathbb{R}^{T_a \times d}\), \(\mathrm{Emb}_a(\cdot)\) denotes the learnable audio-token embedding layer, \(E_{\mathrm{lvl}}^{a}\) is the learnable codebook-level embedding, \(E_{\mathrm{pos}}^{a}\) is the learnable positional embedding, and \(\mathcal{E}_a\) is the audio transformer encoder.
Let \(M_a \in \{0,1\}^{T_a}\) denote the audio padding mask, where \(M_a(j)=1\) indicates a valid audio token and \(M_a(j)=0\) indicates padding.
The token-level audio features are pooled into an audio embedding:
\begin{equation}
\tilde{z}_a = \mathrm{AttnPool}_a(H_a; M_a),
\qquad
z_a = \mathrm{L2Norm}(\tilde{z}_a),
\label{eq:judge_audio_pool}
\end{equation}
where \(\tilde{z}_a,z_a \in \mathbb{R}^{d_o}\).

\vspace{1em}
\textbf{Emotion branch.}
Let \(e^s\) denote the discrete speaker emotion label.
We map it to a learnable embedding and project it into the shared scoring space:
\begin{equation}
h_e = \mathrm{LayerNorm}\!\big(\mathrm{Emb}_e(e^s)\big),
\qquad
\tilde{z}_e = W_e h_e + b_e,
\qquad
z_e = \mathrm{L2Norm}(\tilde{z}_e),
\label{eq:judge_emo_enc}
\end{equation}
where \(\mathrm{Emb}_e(\cdot)\) is the learnable emotion embedding table, \(h_e \in \mathbb{R}^{d}\), \(W_e \in \mathbb{R}^{d_o \times d}\), \(b_e \in \mathbb{R}^{d_o}\), and \(\tilde{z}_e,z_e \in \mathbb{R}^{d_o}\).

\vspace{1em}
\textbf{Fusion branch.}
To unify all available information, we construct one fusion token for each modality in the \(d\)-dimensional hidden space.
Let
$
o \subseteq \{t,a,e\}
$
denote the active modality set, and let \(\delta_k(o)\in\{0,1\}\) indicate whether modality \(k\in\{t,a,e\}\) is available under mode \(o\).

For text and audio, we summarize the hidden states by masked mean pooling over valid positions:
\begin{equation}
\bar{h}_t =
\begin{cases}
\dfrac{\sum_{j=1}^{T_t} M_t(j)\,H_t(j)}
{\sum_{j=1}^{T_t} M_t(j)}, & \delta_t(o)=1, \\[8pt]
0, & \delta_t(o)=0,
\end{cases}
\ \
\bar{h}_a =
\begin{cases}
\dfrac{\sum_{j=1}^{T_a} M_a(j)\,H_a(j)}
{\sum_{j=1}^{T_a} M_a(j)}, & \delta_a(o)=1, \\[8pt]
0, & \delta_a(o)=0,
\end{cases}
\label{eq:judge_masked_mean}
\end{equation}
where \(H_t(j),H_a(j)\in\mathbb{R}^{d}\) denote the \(j\)-th hidden states.
For emotion, which is already represented by a single hidden vector, we define
\begin{equation}
\bar{h}_e =
\begin{cases}
h_e, & \delta_e(o)=1,\\
0, & \delta_e(o)=0.
\end{cases}
\label{eq:judge_emo_hidden}
\end{equation}

We then form the modality-specific fusion tokens
\begin{equation}
u_t = \bar{h}_t + E_{\mathrm{type}}^{t}, \qquad
u_a = \bar{h}_a + E_{\mathrm{type}}^{a}, \qquad
u_e = \bar{h}_e + E_{\mathrm{type}}^{e},
\label{eq:judge_modality_tokens}
\end{equation}
where \(u_t,u_a,u_e \in \mathbb{R}^{d}\), and \(E_{\mathrm{type}}^{t}\), \(E_{\mathrm{type}}^{a}\), and \(E_{\mathrm{type}}^{e}\) are learnable type embeddings.

To explicitly encode which modalities are active, we further introduce a learnable mode embedding \(E_{\mathrm{mode}}(o)\in\mathbb{R}^{d}\).
The initial fusion-token sequence is
\begin{equation}
X_f^{(0)}=
\begin{bmatrix}
u_t + E_{\mathrm{mode}}(o)\\
u_a + E_{\mathrm{mode}}(o)\\
u_e + E_{\mathrm{mode}}(o)
\end{bmatrix}
\in \mathbb{R}^{3\times d}.
\label{eq:judge_fusion_input}
\end{equation}

Since some modalities may be absent, we define a modality-presence mask
\begin{equation}
M_f = \big[\delta_t(o),\,\delta_a(o),\,\delta_e(o)\big] \in \{0,1\}^{3}.
\label{eq:judge_fusion_mask}
\end{equation}
The fusion sequence is processed by a transformer encoder with masking:
\begin{equation}
H_f = \mathcal{E}_f\!\big(X_f^{(0)}; M_f\big),
\qquad
\tilde{z}_f = \mathrm{AttnPool}_f(H_f; M_f),
\qquad
z_f = \mathrm{L2Norm}(\tilde{z}_f),
\label{eq:judge_fused_enc}
\end{equation}
where \(H_f \in \mathbb{R}^{3\times d}\), \(\mathcal{E}_f\) denotes the multimodal fusion transformer, and \(\tilde{z}_f,z_f \in \mathbb{R}^{d_o}\).

\vspace{1em}
\textbf{Motion branch.}
Each candidate listener motion is represented as a motion token sequence \(x_m^l\), obtained using a motion VQ-VAE tokenizer.
We map the motion tokens to embeddings, add positional embeddings, and encode them with a motion transformer:
\begin{equation}
H_m = \mathcal{E}_m\!\big(\mathrm{Emb}_m(x_m^l) + E_{\mathrm{pos}}^{m}\big), \
\tilde{z}_m = \mathrm{AttnPool}_m(H_m; M_m), \
z_m = \mathrm{L2Norm}(\tilde{z}_m),
\label{eq:judge_motion_enc}
\end{equation}
where \(H_m \in \mathbb{R}^{T_m \times d}\), \(M_m \in \{0,1\}^{T_m}\) is the motion padding mask, \(\mathrm{Emb}_m(\cdot)\) is the motion-token embedding layer, \(E_{\mathrm{pos}}^{m}\) is the motion positional embedding, \(\mathcal{E}_m\) is the motion transformer encoder, and \(\tilde{z}_m,z_m \in \mathbb{R}^{d_o}\).

\vspace{1em}
\textbf{Compatibility scoring.}
Let \(\phi(\cdot,\cdot)\) denote the embedding-space compatibility function.
Given a condition embedding \(z \in \mathbb{R}^{d_o}\) and a motion embedding \(z_m \in \mathbb{R}^{d_o}\), we define
\begin{equation}
\phi(z, z_m) = \alpha\, z^\top z_m,
\qquad
\alpha = \exp(\tau),
\label{eq:judge_similarity}
\end{equation}
where \(\tau\) is a learnable temperature parameter and \(\alpha>0\) is the corresponding scaling factor.
Because all score-space embeddings are \(\ell_2\)-normalized, Eq.~\eqref{eq:judge_similarity} is a scaled cosine similarity.

The fused compatibility score is defined as
\begin{equation}
s_\psi(C^s, x_m^l) = \phi(z_f, z_m).
\label{eq:judge_score}
\end{equation}
In addition, we compute auxiliary modality-specific compatibility scores
\begin{equation}
s_\psi^{(k)}(C^s, x_m^l) = \phi(z_k, z_m),
\qquad
k \in \{t,a,e\},
\label{eq:judge_score_single_modality}
\end{equation}
which allow the judge to score candidate motions under partial speaker utterances.

\paragraph{Group-wise contrastive training.}
For each speaker utterance \(C_i^s\), we construct a candidate set
\begin{equation}
\mathcal{U}_i
=
\mathcal{G}(C_i^s)\cup \mathcal{S}(C_i^s)\cup \mathcal{N}(C_i^s),
\label{eq:judge_group_set}
\end{equation}
where \(\mathcal{G}(C_i^s)\), \(\mathcal{S}(C_i^s)\), and \(\mathcal{N}(C_i^s)\) denote the Gold, Silver, and Negative listener motion sets, respectively.
During training, we randomly sample a small number of candidates from each tier and encode them jointly.

To improve robustness to incomplete conditions, we randomly vary the active modality set \(o\) during training.
This encourages the judge to remain reliable under different condition modes, including single-modality settings such as text-only and audio-only.

Let \(\mathcal{P}_i \subseteq \mathcal{U}_i\) denote the positive set associated with \(C_i^s\); in our default setting, \(\mathcal{P}_i=\mathcal{G}(C_i^s)\).
Given a condition embedding \(z_i\) (which can be the fused embedding \(z_f\) or an active modality-specific embedding \(z_t,z_a,z_e\)), we optimize the following group-wise InfoNCE objective:
\begin{equation}
\mathcal{L}_{\mathrm{con}}(z)
=
-\frac{1}{|\mathcal{B}|}
\sum_{i \in \mathcal{B}}
\log
\frac{
\sum\limits_{x \in \mathcal{P}_i}
\exp\!\big(\phi(z_i, z_m(x))\big)
}{
\sum\limits_{x \in \mathcal{U}_i}
\exp\!\big(\phi(z_i, z_m(x))\big)
+
\sum\limits_{b \in \mathcal{B}_{\mathrm{bank}}}
\exp\!\big(\beta\,\phi(z_i, z_m(b))\big)
},
\label{eq:judge_contrastive_general}
\end{equation}
where \(\mathcal{B}\) is the mini-batch, \(\mathcal{B}_{\mathrm{bank}}\) is an auxiliary motion bank providing additional generic negatives, \(z_m(x)\) denotes the motion embedding of candidate \(x\), \(z_m(b)\) denotes the embedding of a motion sampled from the bank, and \(\beta\) controls the contribution of bank negatives.
The motion bank discourages the judge from assigning overly high compatibility scores to generic or template-like motions.

We always apply Eq.~\eqref{eq:judge_contrastive_general} to the fused embedding \(z_f\).
For the modality-specific auxiliary losses, we apply it only to the modalities active under the current mode \(o\):
\begin{equation}
\mathcal{L}_{\mathrm{judge}}
=
\lambda_{\mathrm{f}}\,\mathcal{L}_{\mathrm{con}}(z_f)
+
\sum_{k \in o}\lambda_k\,\mathcal{L}_{\mathrm{con}}(z_k),
\qquad
k\in\{t,a,e\},
\label{eq:judge_contrastive_total}
\end{equation}
where \(\lambda_{\mathrm{f}},\lambda_t,\lambda_a,\lambda_e\) are loss weights to balance different loss terms.

\paragraph{Validation of the multimodal judge network.}
Because the judge network is central to our evaluation protocol, we further verify whether its rankings respect the annotated tier ordering
\(
\mathcal{G}\succ\mathcal{S}\succ\mathcal{N}
\).
For any tier \(\mathcal{A}\in\{\mathcal{G},\mathcal{S},\mathcal{N}\}\), we define its mean judge score under condition \(C^s\) as
\begin{equation}
\bar{s}_{\mathcal{A}}(C^s)
=
\frac{1}{|\mathcal{A}(C^s)|}
\sum_{x \in \mathcal{A}(C^s)}
s_\psi(C^s,x).
\label{eq:judge_mean_score}
\end{equation}
We then report
\textbf{Win(G$>$S)},
\textbf{Win(G$>$N)}, and
\textbf{Win(S$>$N)}, defined as
\begin{equation}
\mathrm{Win}(\mathcal{A}>\mathcal{B})
=
\frac{1}{|\mathcal{D}|}
\sum_{C^s\in\mathcal{D}}
\kappa\!\big(
\bar{s}_{\mathcal{A}}(C^s),
\bar{s}_{\mathcal{B}}(C^s)
\big),
\label{eq:judge_win_eval}
\end{equation}
where \((\mathcal{A},\mathcal{B})\in\{(\mathcal{G},\mathcal{S}),(\mathcal{G},\mathcal{N}),(\mathcal{S},\mathcal{N})\}\), \(\mathcal{D}\) denotes the evaluation set of speaker utterances, and
\begin{equation}
\kappa(u,v)=
\begin{cases}
1, & u>v,\\
0.5, & u=v,\\
0, & u<v.
\end{cases}
\label{eq:kappa_cmp}
\end{equation}

We further report \textbf{MRR(G)}, defined as
\begin{equation}
\mathrm{MRR}(\mathcal{G})
=
\frac{1}{|\mathcal{D}|}
\sum_{C^s\in\mathcal{D}}
\frac{1}{
\min_{x\in\mathcal{G}(C^s)}
\operatorname{rank}_{C^s}(x)
},
\label{eq:judge_mrr}
\end{equation}
where all candidates in \(\mathcal{U}(C^s)\) are sorted in descending order of \(s_\psi(C^s,x)\), and \(\operatorname{rank}_{C^s}(x)\) denotes the resulting 1-based rank of candidate \(x\).

Finally, we report \textbf{nDCG@3}, \textbf{nDCG@5}, and \textbf{nDCG@10}, using graded relevance labels \(2\), \(1\), and \(0\) for Gold, Silver, and Negative candidates, respectively.
These metrics verify whether the learned judge produces rankings aligned with the annotated appropriateness structure.

\paragraph{Strict-L2 missing-modality injection.}
For partial-condition evaluation, we adopt a \emph{Strict-L2} missing-modality injection protocol.
Given an active modality set \(o \subseteq \{t,a,e\}\), every unavailable modality is replaced by a null input \emph{before} it is processed by its encoder branch.
This differs from a weak masking strategy that removes a modality only during fusion while still allowing its encoder to observe the original input.

Formally, let \(\delta_t(o)\), \(\delta_a(o)\), and \(\delta_e(o)\) indicate whether text, audio, and emotion are active under mode \(o\), respectively.
For text, if \(\delta_t(o)=0\), we replace the transcript with an all-padding sequence and set its padding mask to zero:
\begin{equation}
x_t^s \leftarrow \texttt{PAD},
\qquad
M_t(j)=0,\ \forall j.
\end{equation}
For audio, if \(\delta_a(o)=0\), we replace all codec tokens with the audio padding index and mark all time steps as padded:
\begin{equation}
x_a^s \leftarrow \texttt{PAD}_a,
\qquad
M_a(j)=0,\ \forall j.
\end{equation}
For emotion, if \(\delta_e(o)=0\), we replace the original label with a dedicated unknown symbol:
\begin{equation}
e^s \leftarrow \texttt{<unk>}.
\end{equation}
At the fusion stage, the corresponding modality token is additionally masked out through \(M_f\).

As a result, unavailable modalities contribute no semantic information to the final condition representation.
This protocol provides a strict test of whether the judge can reliably score listener motions using only the actually available speaker signals.
Unless otherwise specified, all partial-condition reliability experiments are conducted under this Strict-L2 protocol.

\subsection{Implementation Details of Judge Network}
\label{sec:judge_impl}
\begin{table}[h]
\centering
\caption{Hyperparameters for the multimodal judge network.}
\begin{tabular}{lc}
\toprule
\textbf{Parameter} & \textbf{Value} \\
\midrule
Backbone encoder & T5-base \\
Hidden dimension $d$ & 768 \\
Embedding dimension & 512 \\
Transformer heads & 12 \\
Transformer layers & 6 \\
Feedforward dimension & 3072 \\
Dropout & 0.1 \\
Temperature & 0.07 \\
Memory bank size & 4096 \\
Optimizer & AdamW \\
Learning rate & $5\times10^{-5}$ \\
Weight decay & 0.01 \\
Batch size & 16 \\
Epoch & 50 \\
$\lambda_{\mathrm{f}}$ & 1.0 \\
$\lambda_{\mathrm{t}}$ & 0.5\\
$\lambda_{\mathrm{a}}$ & 0.5 \\
$\lambda_{\mathrm{e}}$ & 0.2 
\\
\bottomrule
\end{tabular}
\label{tab:judge_hyper}
\end{table}

The multimodal judge network is implemented using a transformer-based architecture that evaluates the compatibility between speaker utterances and candidate listener motions.
The textual modality is encoded using a pre-trained T5-base encoder, while audio tokens, emotion labels, and motion tokens are embedded and processed through transformer encoders to obtain modality representations.
These representations are projected into a shared embedding space where the final compatibility score is computed.

Table~\ref{tab:judge_hyper} summarizes the key hyperparameters used for training the judge network.
The model adopts a hidden dimension of 768 and projects the representations into a 512-dimensional embedding space.
The transformer encoder uses 12 attention heads and 6 layers with a feedforward dimension of 3072.
Training is performed using the AdamW optimizer with a learning rate of $5\times10^{-5}$, weight decay of 0.01, and batch size of 16.
A memory bank of size 4096 is used to provide additional negative samples for contrastive training.

\begin{table}[t]
\centering
\setlength{\belowcaptionskip}{5pt}
\caption{
We evaluate the multi-modal matching judge on validation and test set across six input modes (text $T$, audio $A$, emotion $E$, and their fusions).
We report pairwise win rates based on mean score comparisons (Win(G{>}N), Win(G{>}S), Win(S{>}N)) and ranking metrics (MRR(G), nDCG@K with graded relevance G>S>N),
where $\text{G}{=}\mathcal{G}$ (Gold), $\text{S}{=}\mathcal{S}$ (Silver), and $\text{N}{=}\mathcal{N}$ (Negative).
}
\label{tab:judge_metrics_strictL2_supple}
\scriptsize
\setlength{\tabcolsep}{3pt}
\renewcommand{\arraystretch}{1.05}
\resizebox{\linewidth}{!}{%

\begin{tabular}{l l c c c c c c c}
\toprule
\textbf{Mode} & \textbf{Split} &
\textbf{Win(G$>$N)} $\uparrow$ &
\textbf{Win(G$>$S)} $\uparrow$ &
\textbf{Win(S$>$N)} $\uparrow$ &
\textbf{MRR(G)} $\uparrow$ &
\textbf{nDCG@3} $\uparrow$ &
\textbf{nDCG@5} $\uparrow$ &
\textbf{nDCG@10} $\uparrow$ \\
\midrule

\multicolumn{9}{l}{\textbf{Val}} \\
T       & Val  & 0.990 & 0.873 & 0.985 & 0.839 & 0.878 & 0.891 & 0.939 \\
A       & Val  & 0.990 & 0.873 & 0.985 & \textbf{0.842} & \textbf{0.881} & \textbf{0.893} & \textbf{0.940} \\
T+A     & Val  & 0.993 & \textbf{0.883} & \textbf{0.988} & 0.840 & 0.875 & 0.890 & 0.937 \\
T+E     & Val  & \textbf{0.994} & 0.881 & \textbf{0.988} & 0.841 & 0.875 & 0.891 & 0.938 \\
A+E     & Val  & 0.990 & 0.875 & 0.985 & 0.840 & 0.878 & 0.892 & 0.939 \\
T+A+E   & Val  & 0.993 & 0.882 & \textbf{0.988} & 0.840 & 0.876 & 0.890 & 0.937 \\
\midrule

\multicolumn{9}{l}{\textbf{Test}} \\
T       & Test & 0.992 & 0.873 & \textbf{0.983} & 0.829 & 0.864 & \textbf{0.878} & 0.932 \\
A       & Test & 0.992 & 0.872 & \textbf{0.983} & \textbf{0.832} & \textbf{0.866} & \textbf{0.878} & \textbf{0.933} \\
T+A     & Test & \textbf{0.993} & \textbf{0.879} & 0.982 & 0.820 & 0.855 & 0.875 & 0.928 \\
T+E     & Test & \textbf{0.993} & 0.876 & 0.982 & 0.826 & 0.857 & 0.876 & 0.929 \\
A+E     & Test & 0.992 & 0.874 & \textbf{0.983} & 0.831 & 0.865 & \textbf{0.878} & \textbf{0.933} \\
T+A+E   & Test & \textbf{0.993} & 0.878 & 0.982 & 0.828 & 0.859 & \textbf{0.878} & 0.930 \\
\bottomrule
\end{tabular}
}
\end{table}

\subsection{Baseline Methods}
\label{sec:baseline_network}
\paragraph{\textbf{GT.}}
We use the ground-truth listener motion sequences from the test set as an upper-bound reference.

\paragraph{\textbf{Random Selection.}}
We randomly sample a motion sequence from \textsc{HumanML3D}~\cite{guo2022generating} as a naive baseline.

\paragraph{\textbf{Retrieval.}}
Following standard text--motion matching protocols~\cite{wu2025mg,guo2022generating}, we retrieve a listener motion by matching the speaker transcription against candidate motions and returning the top-1 nearest neighbor from the training set.
Specifically, we use the pretrained text and motion encoders from~\cite{guo2022generating}, which are trained with a contrastive objective so that matched text--motion pairs are close in the shared embedding space, while mismatched pairs are separated by a margin.
The text encoder maps the input transcription to a semantic feature vector, while the motion encoder first converts a pose sequence into motion snippet codes and then maps them to a motion feature vector.
In practice, the text encoder follows the architecture in~\cite{guo2022generating}, and the motion encoder is implemented as a bidirectional GRU with hidden size 1,024.

\paragraph{\textbf{Cascaded LLM$\rightarrow$T2M.}}
We construct cascaded baselines by first prompting an LLM to generate the caption of listener reactive motion conditioned on the speaker transcription and emotion.
Then, we feed the generated caption into a text-to-motion (T2M) model to synthesize the final motion.
Here, we consider two LLMs, \textsc{Qwen3-30B-A3B} and a fine-tuned \textsc{Qwen3-4B-Thinking}, together with two representative T2M generators, \textsc{T2M-GPT} and \textsc{MG-MotionLLM}.

Accordingly, \textbf{LLM$\rightarrow$T2M-GPT} denotes the cascade using \textsc{Qwen3-30B-A3B} and \textsc{T2M-GPT}, while \textbf{LLM$\rightarrow$T2M-GPT}$^{*}$ uses the fine-tuned \textsc{Qwen3-4B-Thinking} together with \textsc{T2M-GPT}.
Similarly, \textbf{LLM$\rightarrow$MG-MotionLLM} denotes the cascade using \textsc{Qwen3-30B-A3B} and \textsc{MG-MotionLLM}, while \textbf{LLM$\rightarrow$MG-MotionLLM}$^{*}$ uses the fine-tuned \textsc{Qwen3-4B-Thinking} together with \textsc{MG-MotionLLM}.

To keep the main table concise, we report the cascaded baselines under the $T{+}E$ setting.

\subsection{Evaluation Metrics}
\label{sec:supp_metrics}

We evaluate model performance from three complementary perspectives:
(i) \emph{reactive appropriateness},
(ii) \emph{motion quality}, and
(iii) \emph{diversity}.

\paragraph{Reactive appropriateness.}
Reactive appropriateness measures how well the generated listener motions respond to the speaker utterance.
For each speaker utterance \(C^s\), the annotated listener motions are partitioned into three relevance tiers:
Gold \(\mathcal{G}(C^s)\), Silver \(\mathcal{S}(C^s)\), and Negative \(\mathcal{N}(C^s)\).
Let
\begin{equation}
\widehat{\mathcal{R}}^l(C^s)
=
\{\hat{x}_{m,1}^l,\ldots,\hat{x}_{m,M}^l\}
\end{equation}
denote the set of \(M\) generated listener motion sequences for the same condition.
To assess relative appropriateness, we use the multimodal judge network introduced in Sec.~\ref{sec:judge_network}, which assigns a compatibility score
\begin{equation}
s_\psi(C^s, x_m^l)
\end{equation}
to a candidate listener motion \(x_m^l\) conditioned on the speaker input \(C^s\).

For any candidate set \(\mathcal{A}(C^s)\), we define its mean judge score as
\begin{equation}
\bar{s}_{\mathcal{A}}(C^s)
=
\frac{1}{|\mathcal{A}(C^s)|}
\sum_{x_m^l \in \mathcal{A}(C^s)}
s_\psi(C^s, x_m^l).
\label{eq:mean_judge_score_eval}
\end{equation}
For brevity, we denote the mean scores of the generated set and the three annotated tiers by
\begin{equation}
g(C^s)=\bar{s}_{\widehat{\mathcal{R}}^l}(C^s), \
G(C^s)=\bar{s}_{\mathcal{G}}(C^s), \
S(C^s)=\bar{s}_{\mathcal{S}}(C^s), \
N(C^s)=\bar{s}_{\mathcal{N}}(C^s).
\label{eq:mean_scores_gsn}
\end{equation}

We then report
\textbf{Win(g$>$G)},
\textbf{Win(g$>$S)}, and
\textbf{Win(g$>$N)}, defined as
\begin{equation}
\mathrm{Win}(g>\mathcal{A})
=
\frac{1}{|\mathcal{D}|}
\sum_{C^s \in \mathcal{D}}
\kappa\!\big(
g(C^s), \bar{s}_{\mathcal{A}}(C^s)
\big),
\qquad
\mathcal{A}\in\{\mathcal{G},\mathcal{S},\mathcal{N}\},
\label{eq:win_gen_vs_ref}
\end{equation}
where \(\mathcal{D}\) denotes the evaluation set, and
\begin{equation}
\kappa(u,v)=
\begin{cases}
1, & u>v,\\
0.5, & u=v,\\
0, & u<v.
\end{cases}
\label{eq:kappa_cmp_1}
\end{equation}
Intuitively, \textbf{Win(g$>$N)} measures whether the generated motions are preferred over clearly inappropriate responses,
\textbf{Win(g$>$S)} is a stricter criterion against moderately appropriate responses,
and \textbf{Win(g$>$G)} is the most challenging criterion against highly appropriate annotated reactions.
Higher values indicate stronger reactive appropriateness.

We further report \textbf{Gen@3}, which measures whether at least one generated motion is ranked within the top 3 among all candidates under the same speaker utterance.
For each \(C^s\), we form the candidate pool
\begin{equation}
\mathcal{C}(C^s)
=
\mathcal{G}(C^s)\cup
\mathcal{S}(C^s)\cup
\mathcal{N}(C^s)\cup
\widehat{\mathcal{R}}^l(C^s),
\label{eq:candidate_pool_eval}
\end{equation}
rank all candidates in \(\mathcal{C}(C^s)\) by \(s_\psi(C^s,\cdot)\) in descending order, and denote the resulting rank of a candidate \(x_m^l\) by \(\operatorname{rank}_{C^s}(x_m^l)\).
We then compute
\begin{equation}
\mathrm{Gen@3}
=
\frac{1}{|\mathcal{D}|}
\sum_{C^s \in \mathcal{D}}
\mathbb{I}\!\left[
\min_{\hat{x}_m^l \in \widehat{\mathcal{R}}^l(C^s)}
\operatorname{rank}_{C^s}(\hat{x}_m^l)\le 3
\right].
\label{eq:gen_at3_eval}
\end{equation}
This metric is particularly suitable for our task because reactive listener behavior is inherently one-to-many: the same speaker utterance may admit multiple plausible listener reactions, and \textbf{Gen@3} evaluates whether the model can produce at least one highly competitive response within a limited candidate budget.

\paragraph{Motion quality.}
We evaluate motion quality using Fr\'echet Inception Distance (FID)~\cite{FID} in a motion feature space.
Let \(f_{\mathrm{eval}}(x_m^l)\) denote the feature representation of a motion sequence extracted by a pretrained motion evaluation network.
We compute the feature statistics of generated motions and real motions in the test set, and then measure the Fr\'echet distance between the two Gaussian distributions:
\begin{equation}
\mathrm{FID}
=
\|\mu_r-\mu_g\|_2^2
+
\operatorname{Tr}\!\left(
\Sigma_r+\Sigma_g-2(\Sigma_r\Sigma_g)^{1/2}
\right),
\label{eq:fid_eval}
\end{equation}
where \((\mu_r,\Sigma_r)\) and \((\mu_g,\Sigma_g)\) are the mean and covariance of the real and generated motion features, respectively.
Lower FID indicates that the generated motions are closer to the distribution of real listener motions, and therefore reflects better overall motion quality.

\paragraph{Diversity.}
Since a single speaker utterance may admit multiple plausible listener reactions, it is also important to evaluate the diversity of generated motions.
Following prior work in human motion generation~\cite{wu2025mg,zhang2023generating}, we measure diversity in the same motion feature space.
Given the set of all generated motions, we randomly sample two subsets of equal size \(S_d\), denoted by
\(\{\hat{x}_{m,1}^l,\dots,\hat{x}_{m,S_d}^l\}\) and
\(\{\hat{x}_{m,1}^{l\prime},\dots,\hat{x}_{m,S_d}^{l\prime}\}\),
and define diversity as
\begin{equation}
\mathrm{Diversity}
=
\frac{1}{S_d}
\sum_{i=1}^{S_d}
\left\|
f_{\mathrm{eval}}(\hat{x}_{m,i}^l)
-
f_{\mathrm{eval}}(\hat{x}_{m,i}^{l\prime})
\right\|_2.
\label{eq:diversity_eval}
\end{equation}
Higher diversity indicates that the generated motions exhibit greater variation and are less likely to collapse to a small set of repetitive motion patterns.

\setlength{\LTleft}{0pt}
\setlength{\LTright}{0pt}
\setlength{\tabcolsep}{3pt}  
\renewcommand{\arraystretch}{1.0}
{\small
\begin{longtable}{@{}ccccccccc@{}}
\caption{Full hyperparameter sweep results for group-wise preference training. We vary the ranking margin $m$, ranking-loss weight $\lambda_{\mathrm{rank}}$, and Gold-vs-Negative weight $\lambda_{\mathrm{gn}}$. We report pairwise preference metrics (Win(g$>$N), Win(g$>$S), Win(g$>$G)), together with Gen@3, FID, and Diversity.}
\label{tab:supp_hparam_full}\\
\toprule
$m$ & $\lambda_{\mathrm{rank}}$ & $\lambda_{\mathrm{gn}}$
& \makecell{Win(g$>$N) $\uparrow$ }
& \makecell{Win(g$>$S) $\uparrow$ }
& \makecell{Win(g$>$G) $\uparrow$ }
& \makecell{Gen@3 $\uparrow$ }
& \makecell{FID $\downarrow$}
& \makecell{Diversity $\uparrow$ } \\
\midrule
\endfirsthead
\toprule
$m$ & $\lambda_{\mathrm{rank}}$ & $\lambda_{\mathrm{gn}}$
& \makecell{Win(g$>$N) $\uparrow$ }
& \makecell{Win(g$>$S) $\uparrow$ }
& \makecell{Win(g$>$G) $\uparrow$ }
& \makecell{Gen@3 $\uparrow$ }
& \makecell{FID $\downarrow$}
& \makecell{Diversity $\uparrow$ } \\
\midrule
\endhead

\bottomrule
\endfoot
0.00 & 0.00 & 0.00 & 0.9976 & 0.7809 & 0.2585 & 0.9600 & 5.2638 & 5.3005 \\
0.00 & 0.00 & 0.25 & 0.9976 & 0.7809 & 0.2633 & 0.9600 & 5.2638 & 5.3005 \\
0.00 & 0.00 & 0.50 & 0.9976 & 0.7809 & 0.2615 & 0.9600 & 5.2638 & 5.3005 \\
0.00 & 0.00 & 1.00 & 0.9964 & 0.7809 & 0.2615 & 0.9613 & 5.2638 & 5.3005 \\
0.00 & 0.25 & 0.00 & 0.9988 & 0.7809 & 0.2331 & 0.9467 & 5.9644 & 4.6993 \\
0.00 & 0.25 & 0.25 & 0.9952 & 0.7482 & 0.2240 & 0.9467 & 5.2102 & 4.8197 \\
0.00 & 0.25 & 0.50 & 0.9988 & 0.7288 & 0.2137 & 0.9455 & 5.3426 & 4.9865 \\
0.00 & 0.25 & 1.00 & 0.9939 & 0.7815 & 0.2458 & 0.9528 & 5.3948 & 4.7384 \\
0.00 & 0.50 & 0.00 & 0.9927 & 0.7760 & 0.2548 & 0.9443 & 4.6552 & 4.7315 \\
0.00 & 0.50 & 0.25 & 0.9952 & 0.7730 & 0.2482 & 0.9600 & 5.4479 & 4.4127 \\
0.00 & 0.50 & 0.50 & 0.9952 & 0.7476 & 0.2379 & 0.9600 & 5.9814 & 4.3124 \\
0.00 & 0.50 & 1.00 & 0.9939 & 0.7694 & 0.2512 & 0.9576 & 5.3426 & 4.5137 \\
0.00 & 1.00 & 0.00 & 0.9964 & 0.7548 & 0.2312 & 0.9443 & 6.5379 & 3.9613 \\
0.00 & 1.00 & 0.25 & 0.9891 & 0.7306 & 0.2391 & 0.9479 & 7.0065 & 3.9543 \\
0.00 & 1.00 & 0.50 & 0.9964 & 0.7391 & 0.2125 & 0.9540 & 5.5322 & 4.4312 \\
0.00 & 1.00 & 1.00 & 0.9855 & 0.6731 & 0.1925 & 0.9407 & 6.8036 & 3.9632 \\
0.50 & 0.00 & 0.00 & 0.9964 & 0.7809 & 0.2597 & 0.9600 & 5.2638 & 5.3005 \\
0.50 & 0.00 & 0.25 & 0.9976 & 0.7809 & 0.2639 & 0.9613 & 5.2638 & 5.3005 \\
0.50 & 0.00 & 0.50 & 0.9976 & 0.7809 & 0.2615 & 0.9600 & 5.2638 & 5.3005 \\
0.50 & 0.00 & 1.00 & 0.9952 & 0.7809 & 0.2615 & 0.9588 & 5.2638 & 5.3005 \\
0.50 & 0.25 & 0.00 & 0.9939 & 0.7494 & 0.2349 & 0.9407 & 5.0807 & 4.8318 \\
0.50 & 0.25 & 0.25 & 1.0000 & 0.7966 & 0.2663 & 0.9600 & 4.7596 & 4.8039 \\
0.50 & 0.25 & 0.50 & 0.9903 & 0.7337 & 0.2343 & 0.9407 & 4.8888 & 4.6845 \\
0.50 & 0.25 & 1.00 & 0.9952 & 0.8184 & 0.2778 & 0.9552 & 5.1955 & 4.8183 \\
0.50 & 0.50 & 0.00 & 0.9964 & 0.8287 & 0.3057 & 0.9625 & 5.8396 & 4.1884 \\
0.50 & 0.50 & 0.25 & 0.9952 & 0.7579 & 0.2318 & 0.9310 & 5.3855 & 4.3443 \\
0.50 & 0.50 & 0.50 & 0.9952 & 0.7736 & 0.2385 & 0.9625 & 6.2371 & 4.3488 \\
0.50 & 0.50 & 1.00 & 0.9952 & 0.6762 & 0.1913 & 0.9467 & 6.1306 & 4.3766 \\
0.50 & 1.00 & 0.00 & 0.9915 & 0.7337 & 0.2403 & 0.9492 & 6.7096 & 3.9289 \\
0.50 & 1.00 & 0.25 & 0.9915 & 0.7082 & 0.2149 & 0.9443 & 5.4811 & 4.1878 \\
0.50 & 1.00 & 0.50 & 0.9673 & 0.6132 & 0.1901 & 0.9334 & 6.9334 & 3.9102 \\
0.50 & 1.00 & 1.00 & 0.9891 & 0.6168 & 0.1834 & 0.9237 & 6.5986 & 3.9541 \\
1.00 & 0.00 & 0.00 & 0.9976 & 0.7809 & 0.2597 & 0.9600 & 5.2638 & 5.3005 \\
1.00 & 0.00 & 0.25 & 0.9976 & 0.7809 & 0.2609 & 0.9588 & 5.2638 & 5.3005 \\
1.00 & 0.00 & 0.50 & 0.9964 & 0.7809 & 0.2609 & 0.9600 & 5.2638 & 5.3005 \\
1.00 & 0.00 & 1.00 & 0.9976 & 0.7809 & 0.2627 & 0.9588 & 5.2638 & 5.3005 \\
1.00 & 0.25 & 0.00 & 0.9964 & 0.8008 & 0.2851 & 0.9516 & 6.0285 & 4.2946 \\
1.00 & 0.25 & 0.25 & 0.9939 & 0.7676 & 0.2464 & 0.9552 & 5.1537 & 4.6242 \\
1.00 & 0.25 & 0.50 & 0.9939 & 0.7821 & 0.2682 & 0.9516 & 5.3639 & 4.5391 \\
1.00 & 0.25 & 1.00 & 0.9988 & 0.8117 & 0.2706 & 0.9625 & 5.1943 & 4.6935 \\
1.00 & 0.50 & 0.00 & 0.9927 & 0.7524 & 0.2288 & 0.9528 & 5.3754 & 4.3702 \\
1.00 & 0.50 & 0.25 & 0.9952 & 0.7361 & 0.2288 & 0.9455 & 5.6698 & 4.2394 \\
1.00 & 0.50 & 0.50 & 0.9903 & 0.7113 & 0.2010 & 0.9516 & 5.8942 & 4.3384 \\
1.00 & 0.50 & 1.00 & 0.9915 & 0.6501 & 0.1816 & 0.9310 & 5.6888 & 4.2328 \\
1.00 & 1.00 & 0.00 & 0.9952 & 0.6562 & 0.1973 & 0.9310 & 7.0648 & 3.9867 \\
1.00 & 1.00 & 0.25 & 0.9849 & 0.5938 & 0.1774 & 0.9262 & 7.4283 & 3.8852 \\
1.00 & 1.00 & 0.50 & 0.9921 & 0.5914 & 0.1798 & 0.9104 & 8.6083 & 3.6349 \\
1.00 & 1.00 & 1.00 & 0.9831 & 0.5847 & 0.1731 & 0.9237 & 6.2941 & 3.9609 \\
2.00 & 0.00 & 0.00 & 0.9976 & 0.7809 & 0.2567 & 0.9600 & 5.2638 & 5.3005 \\
2.00 & 0.00 & 0.25 & 0.9976 & 0.7809 & 0.2585 & 0.9600 & 5.2638 & 5.3005 \\
2.00 & 0.00 & 0.50 & 0.9964 & 0.7809 & 0.2579 & 0.9600 & 5.2638 & 5.3005 \\
2.00 & 0.00 & 1.00 & 0.9976 & 0.7809 & 0.2627 & 0.9613 & 5.2638 & 5.3005 \\
2.00 & 0.25 & 0.00 & 0.9952 & 0.7639 & 0.2512 & 0.9540 & 5.6781 & 4.4907 \\
2.00 & 0.25 & 0.25 & 0.9891 & 0.7433 & 0.2452 & 0.9588 & 5.1178 & 4.7459 \\
2.00 & 0.25 & 0.50 & 0.9964 & 0.7815 & 0.2603 & 0.9588 & 5.6664 & 4.3494 \\
2.00 & 0.25 & 1.00 & 0.9939 & 0.7748 & 0.2785 & 0.9697 & 5.7083 & 4.1561 \\
2.00 & 0.50 & 0.00 & 0.9939 & 0.7264 & 0.2228 & 0.9516 & 6.1482 & 4.1211 \\
2.00 & 0.50 & 0.25 & 0.9964 & 0.6477 & 0.1828 & 0.9249 & 6.7075 & 3.8914 \\
2.00 & 0.50 & 0.50 & 0.9964 & 0.6326 & 0.1901 & 0.9249 & 5.4215 & 4.1601 \\
2.00 & 0.50 & 1.00 & 0.9909 & 0.6610 & 0.1907 & 0.9370 & 6.8355 & 3.7096 \\
2.00 & 1.00 & 0.00 & 0.9927 & 0.6423 & 0.1998 & 0.9298 & 7.1093 & 3.8085 \\
2.00 & 1.00 & 0.25 & 0.9715 & 0.6483 & 0.2046 & 0.9407 & 6.8560 & 3.7436 \\
2.00 & 1.00 & 0.50 & 0.9752 & 0.6362 & 0.1907 & 0.9298 & 6.1279 & 3.8659 \\
2.00 & 1.00 & 1.00 & 0.9655 & 0.5648 & 0.1544 & 0.9140 & 6.1125 & 4.0394 \\
\end{longtable}
}

\section{More Details of ReactMotionNet Dataset}
\label{sec:supp_dataset}

\begin{figure}[t]
    \centering

    \begin{minipage}{0.48\linewidth}
        \centering
        \includegraphics[width=\linewidth]{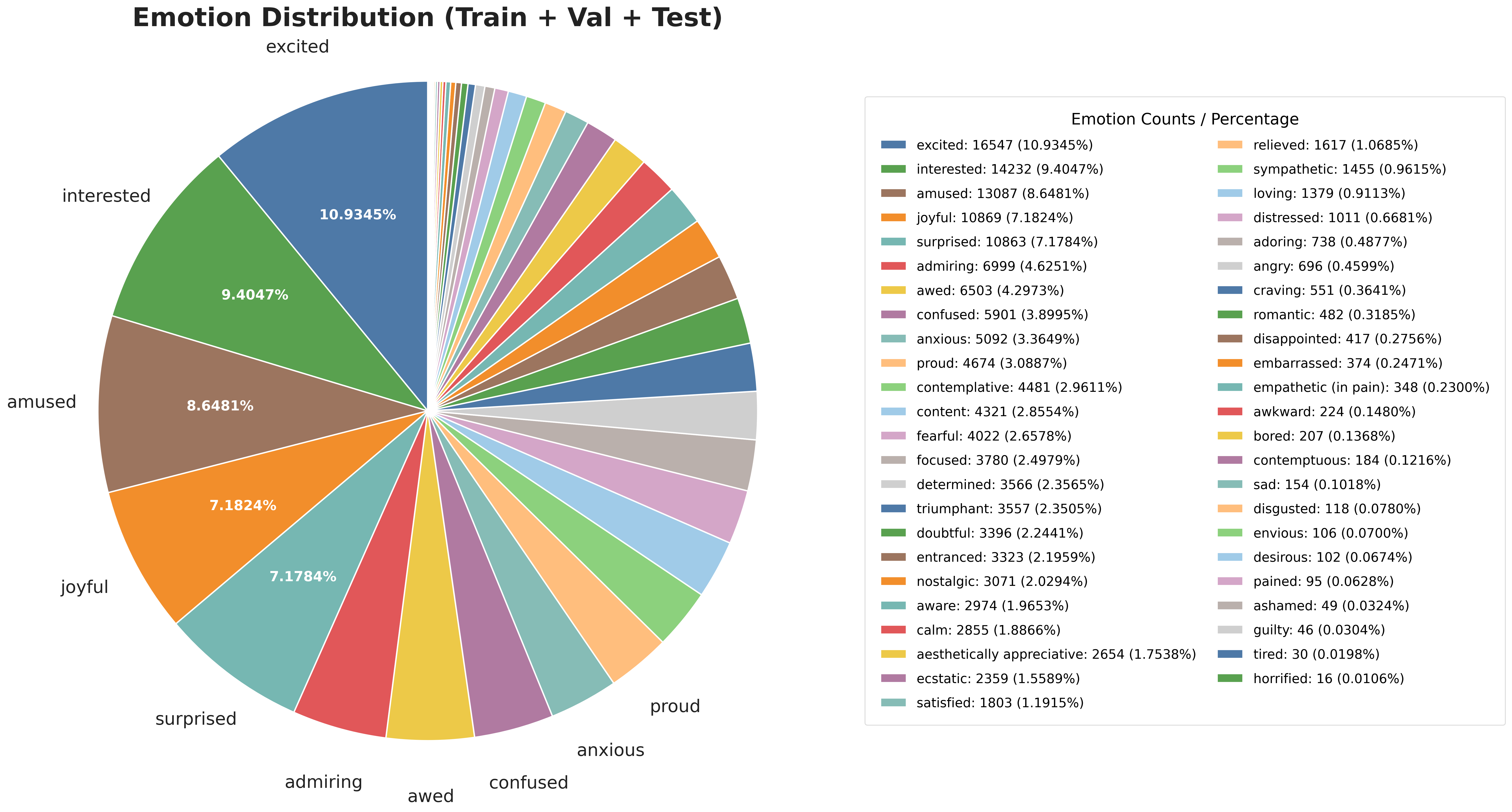}\\
        \small (a) All
    \end{minipage}\hfill
    \begin{minipage}{0.48\linewidth}
        \centering
        \includegraphics[width=\linewidth]{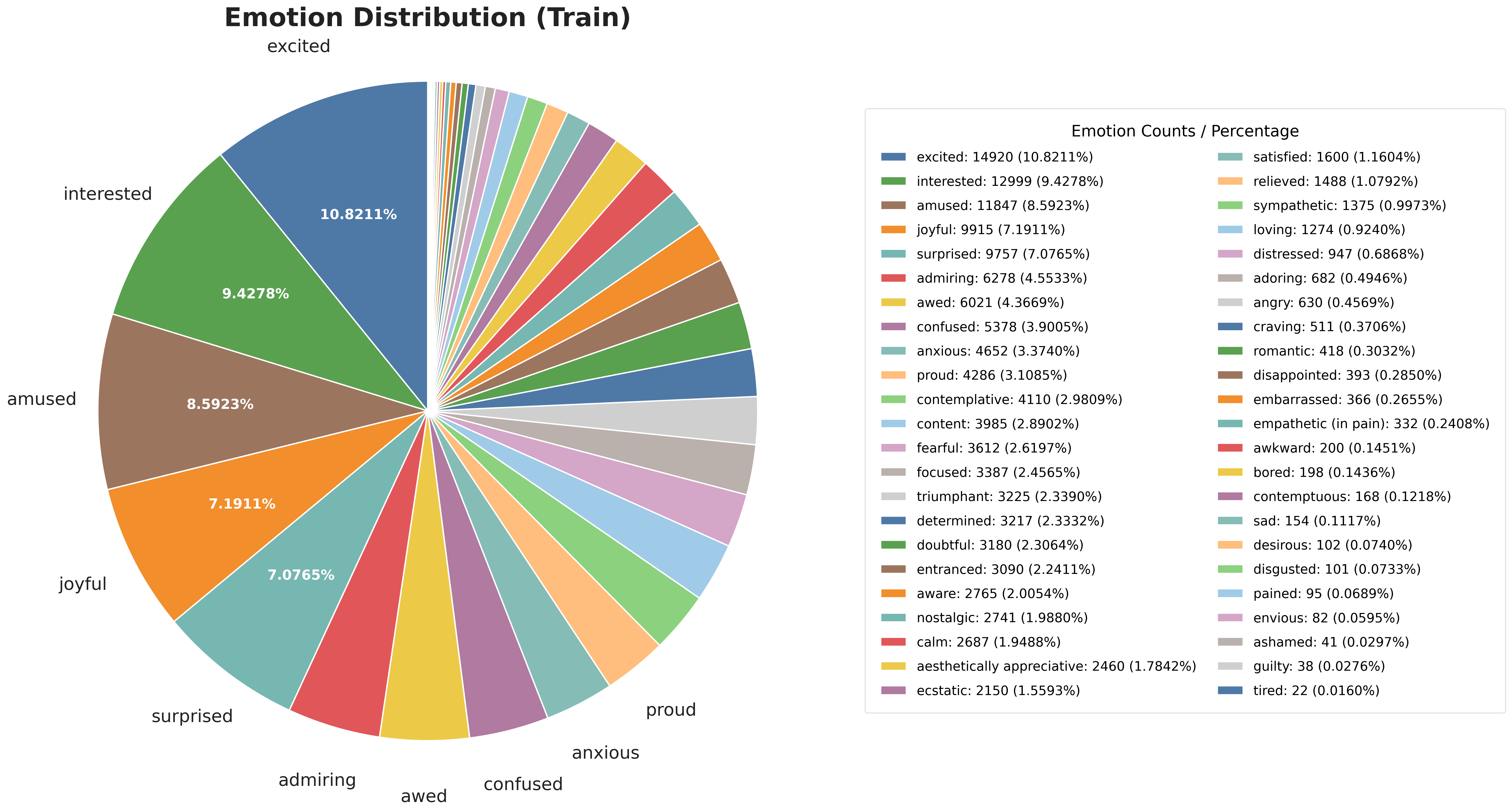}\\
        \small (b) Train
    \end{minipage}

    \vspace{0.5em}

    \begin{minipage}{0.48\linewidth}
        \centering
        \includegraphics[width=\linewidth]{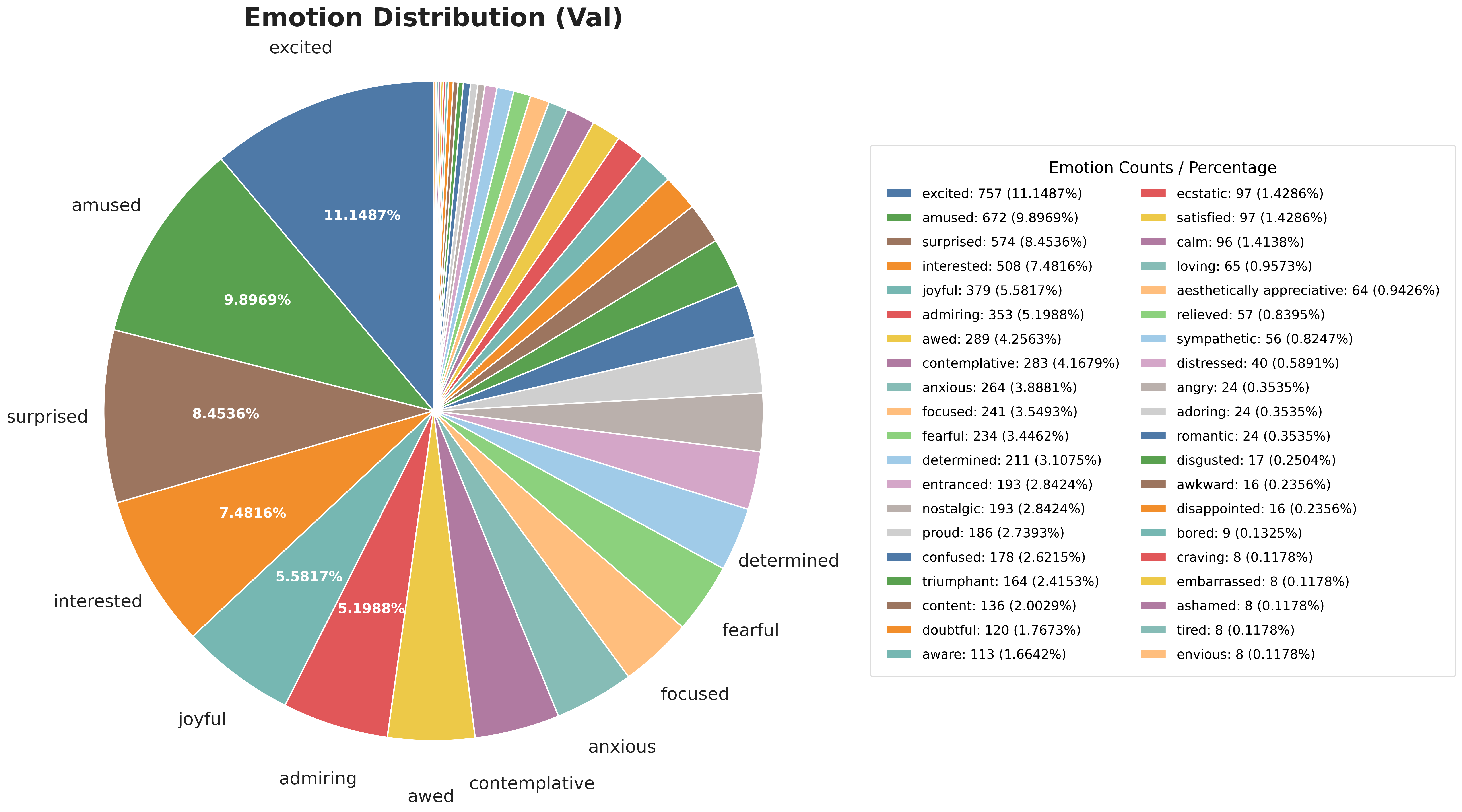}\\
        \small (c) Val
    \end{minipage}\hfill
    \begin{minipage}{0.48\linewidth}
        \centering
        \includegraphics[width=\linewidth]{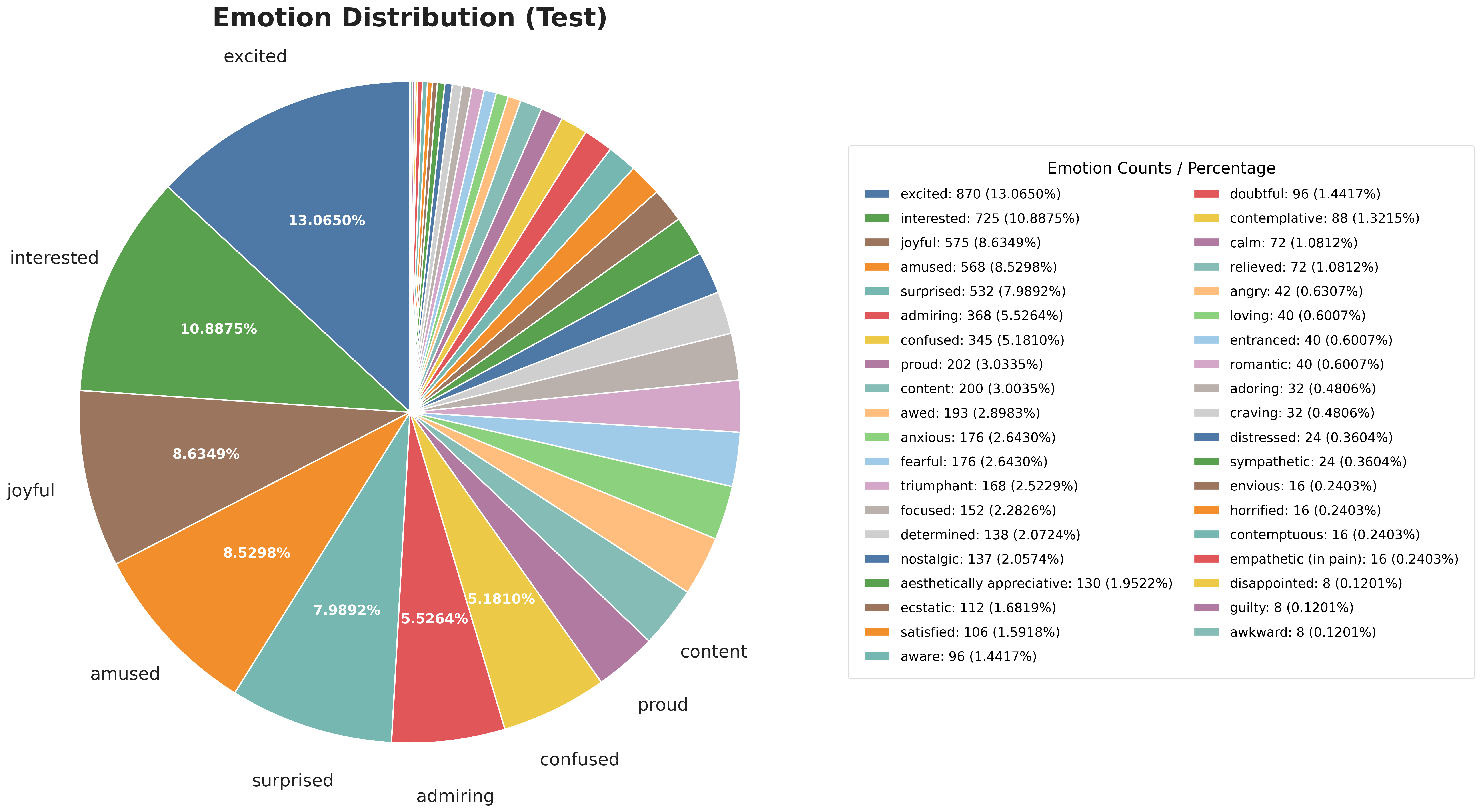}\\
        \small (d) Test
    \end{minipage}

    \caption{Emotion distributions over the full dataset and across the train/validation/test splits.}
    \label{fig:emotion_pies}
\end{figure}

ReactMotionNet exhibits three desirable properties for studying reactive listener motion generation.
First, it provides \emph{large-scale} supervision, containing over 151K labeled speaker--listener pairs.
Second, it explicitly captures the \emph{one-to-many} nature of listener behavior by associating each speaker utterance with multiple candidate reactive motions.
Third, it provides \emph{graded supervision} through Gold, Silver, and Negative labels, supporting both generative modeling and preference-aware evaluation.
Moreover, the dataset is split by disjoint speaker utterances, enabling a cleaner evaluation of generalization to unseen conversational conditions.

In total, ReactMotionNet contains 151{,}328 labeled speaker--listener pairs, covering 8{,}298 unique speaker utterances and 2{,}029 unique listener reactive motions.
On average, each speaker utterance is paired with 18.24 candidate reactive motions, further highlighting the inherently \emph{one-to-many} nature of reactive listener behavior.
Among all pairs, 9{,}307, 34{,}196, and 107{,}825 are annotated as Gold, Silver, and Negative, respectively, reflecting the \emph{graded appropriateness} of candidate reactions.
We partition the dataset by \emph{speaker utterance} using an 8:1:1 train/validation/test split, ensuring that utterances are \emph{disjoint} across splits, \ie, no utterance appears in more than one partition.

The dataset covers 47 emotion categories, including \textit{admiring}, \textit{adoring}, \textit{aesthetically appreciative}, \textit{amused}, \textit{angry}, \textit{anxious}, \textit{ashamed}, \textit{aware}, \textit{awed}, \textit{awkward}, \textit{bored}, \textit{calm}, \textit{confused}, \textit{contemplative}, \textit{contemptuous}, \textit{content}, \textit{craving}, \textit{desirous}, \textit{determined}, \textit{disappointed}, \textit{disgusted}, \textit{distressed}, \textit{doubtful}, \textit{ecstatic}, \textit{embarrassed}, \textit{empathetic (in pain)}, \textit{entranced}, \textit{envious}, \textit{excited}, \textit{fearful}, \textit{focused}, \textit{guilty}, \textit{horrified}, \textit{interested}, \textit{joyful}, \textit{loving}, \textit{nostalgic}, \textit{pained}, \textit{proud}, \textit{relieved}, \textit{romantic}, \textit{sad}, \textit{satisfied}, \textit{surprised}, \textit{sympathetic}, \textit{tired}, and \textit{triumphant}.
As shown in Fig.~\ref{fig:emotion_pies}, these emotion labels exhibit a broad yet imbalanced distribution across the full dataset and each split, making ReactMotionNet a realistic benchmark for modeling diverse affective conversational responses.

\section{Additional Experimental Results}
\label{sec:experiment}
\subsection{Hyperparameter Sensitivity Analysis}
\label{sec:supp_hparam}

We study the sensitivity of group-wise preference training to the ranking margin $m$, the ranking-loss weight $\lambda_{\mathrm{rank}}$, and the Gold-vs-Negative weight $\lambda_{\mathrm{gn}}$. 
We primarily consider \textbf{Gen@3}, which measures whether generated motions can be ranked among the top plausible candidates under the same candidate budget. 
We additionally report \textbf{Win(g$>$S)} and \textbf{Win(g$>$G)} to assess relative preference quality against medium-quality and high-quality reference candidates, respectively. 
FID and Diversity are further included to characterize motion realism and output diversity.

The hyperparameter sweep reveals several consistent patterns. 
First, introducing a \emph{small positive ranking margin} is beneficial and more reliable than using no margin. 
Under $\lambda_{\mathrm{rank}}=0.25$ and $\lambda_{\mathrm{gn}}=0.25$, increasing $m$ from $0$ to $0.5$ improves Win(g$>$S) from $0.7482$ to $0.7966$, Win(g$>$G) from $0.2240$ to $0.2663$, and Gen@3 from $0.9467$ to $0.9600$, while simultaneously reducing FID from $5.2102$ to $4.7596$. 
Although larger margins can further increase Gen@3 in certain cases, such gains are not consistently accompanied by improvements in preference alignment or motion quality, suggesting that excessively large margins may over-specialize the objective.

Second, $\lambda_{\mathrm{rank}}$ is the most sensitive hyperparameter in the sweep. 
Moderate ranking supervision is beneficial, whereas overly large values tend to degrade both alignment and generation quality. 
For instance, at $m=0.5$ and $\lambda_{\mathrm{gn}}=0.25$, increasing $\lambda_{\mathrm{rank}}$ from $0.25$ to $0.5$ and $0.1$ decreases Win(g$>$S) from $0.7966$ to $0.7579$ and $0.7082$, decreases Win(g$>$G) from $0.2663$ to $0.2318$ and $0.2149$, and worsens FID from $4.7596$ to $5.3855$ and $5.4811$. 
This indicates that excessive ranking pressure can bias optimization toward relative ordering at the expense of generative fidelity.

Third, $\lambda_{\mathrm{gn}}$ has a secondary but non-negligible effect, with a \emph{moderate} value yielding the most favorable trade-off. 
At $m=0.5$ and $\lambda_{\mathrm{rank}}=0.25$, setting $\lambda_{\mathrm{gn}}=0.25$ improves Win(g$>$S), Win(g$>$G), and Gen@3 over $\lambda_{\mathrm{gn}}=0$, while also reducing FID. 
By contrast, further increasing $\lambda_{\mathrm{gn}}$ to $1.0$ slightly improves pairwise preference scores, but lowers Gen@3 and degrades FID, indicating that stronger Gold-vs-Negative separation does not necessarily translate into better overall generation quality.

Accordingly, we use $m=0.5$, $\lambda_{\mathrm{rank}}=0.25$, and $\lambda_{\mathrm{gn}}=0.25$ in all main experiments, as this setting resides in a stable regime of the sweep and yields the most balanced overall performance across preference-oriented and generation-oriented criteria.

\begin{figure*}[t]
\centering
\includegraphics[width=0.3\linewidth]{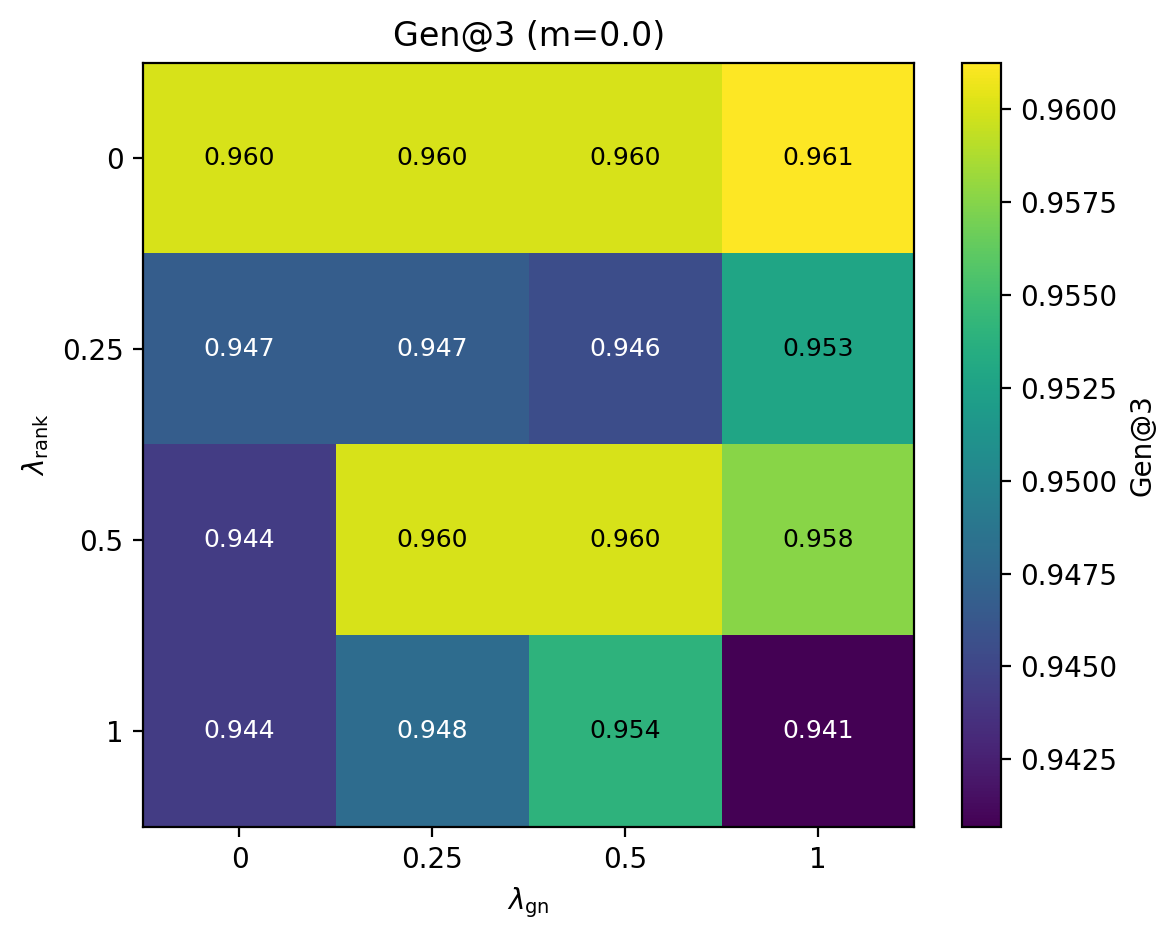}
\includegraphics[width=0.3\linewidth]{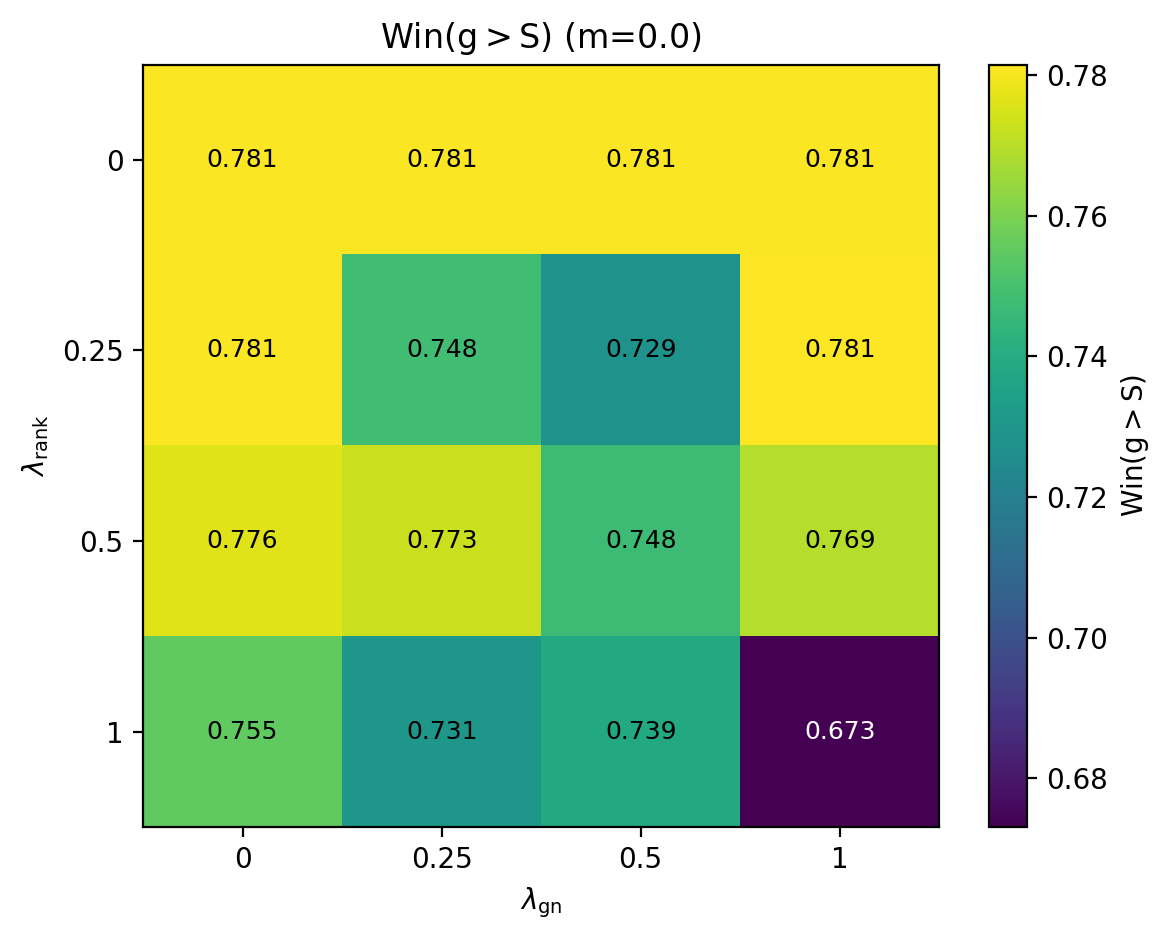}
\includegraphics[width=0.3\linewidth]{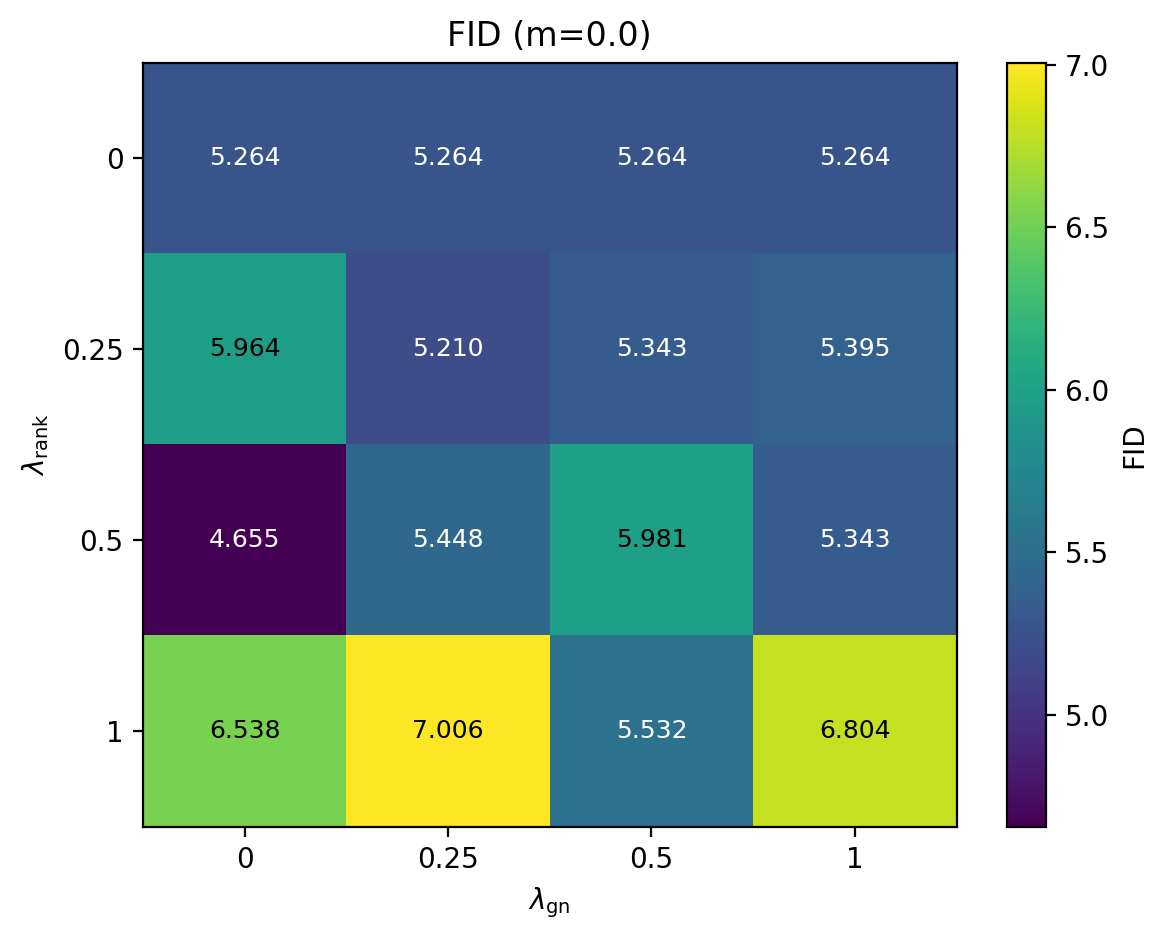}
\\
\includegraphics[width=0.3\linewidth]{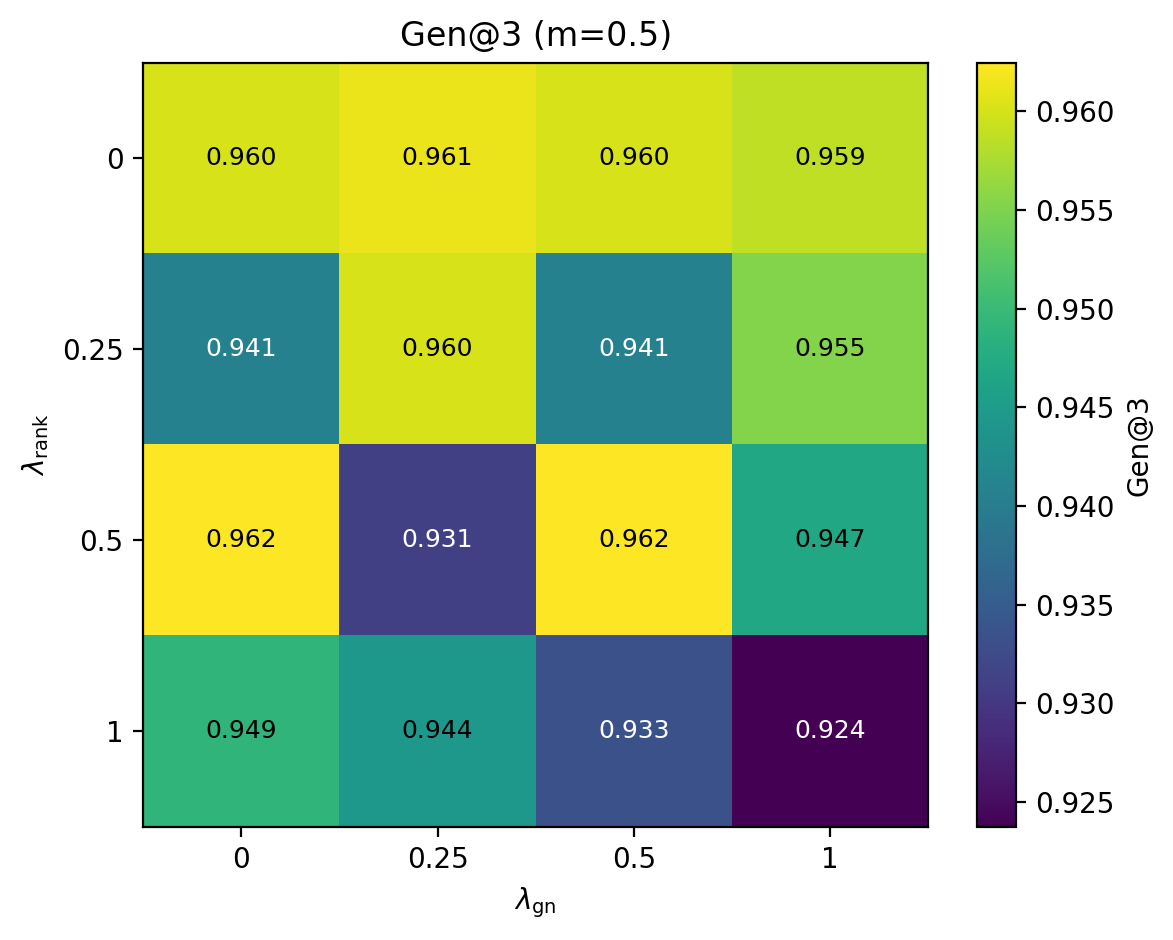}
\includegraphics[width=0.3\linewidth]{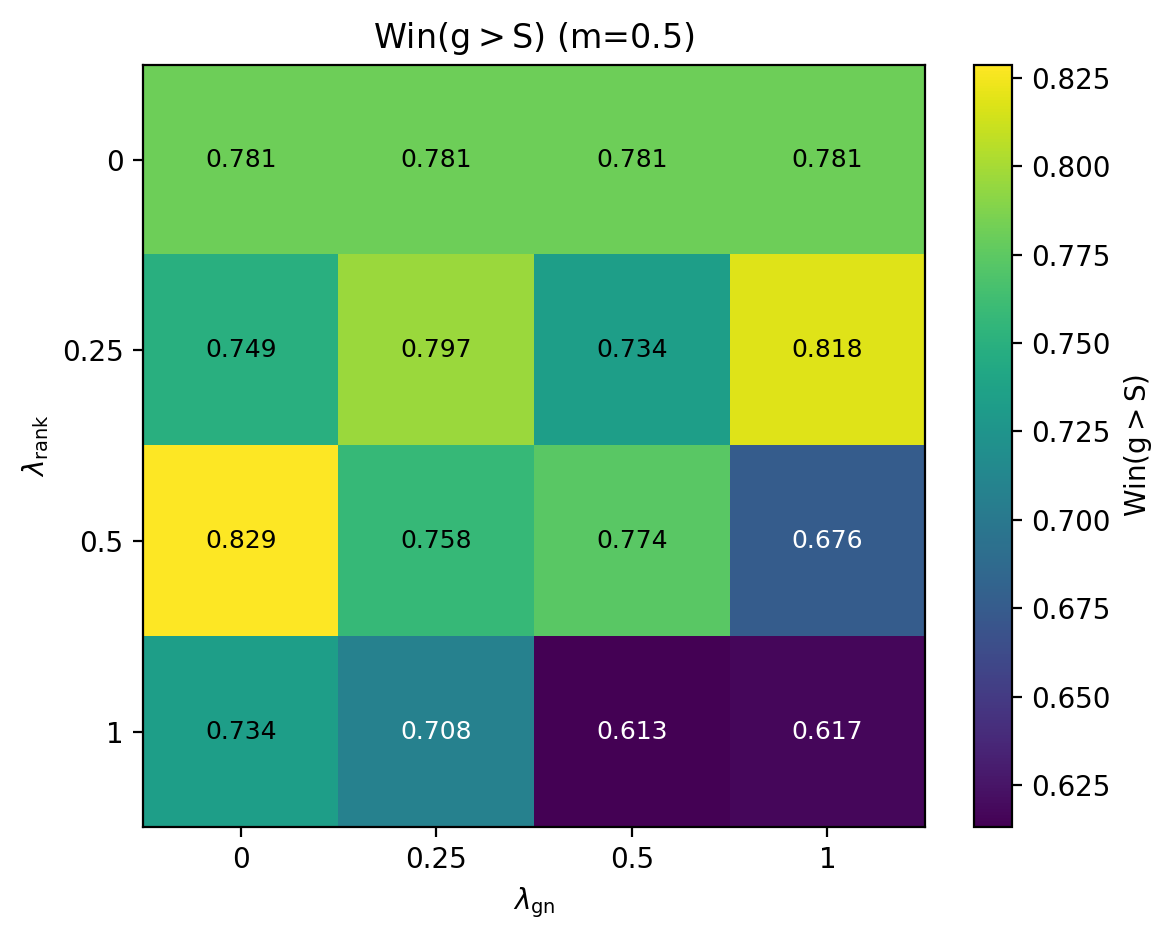}
\includegraphics[width=0.3\linewidth]{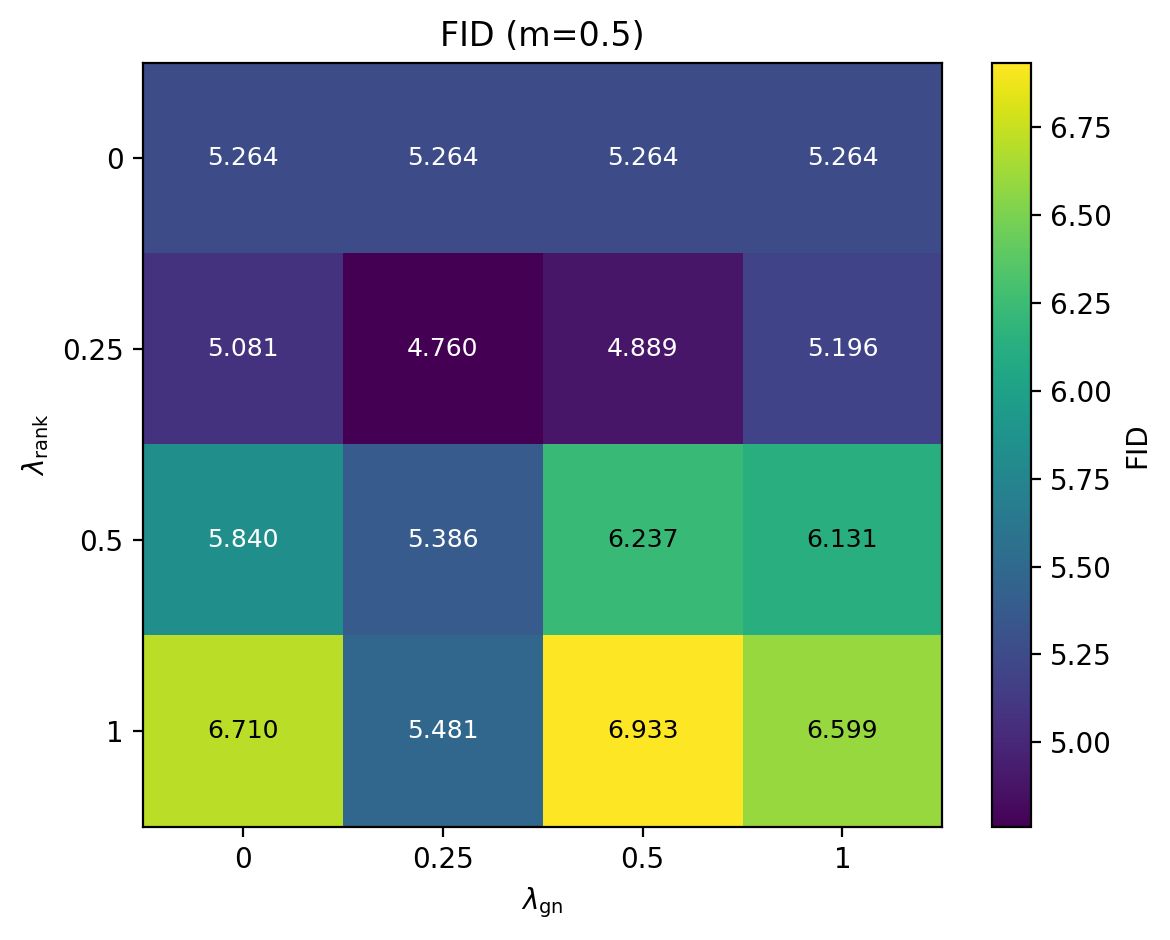}
\\
\includegraphics[width=0.3\linewidth]{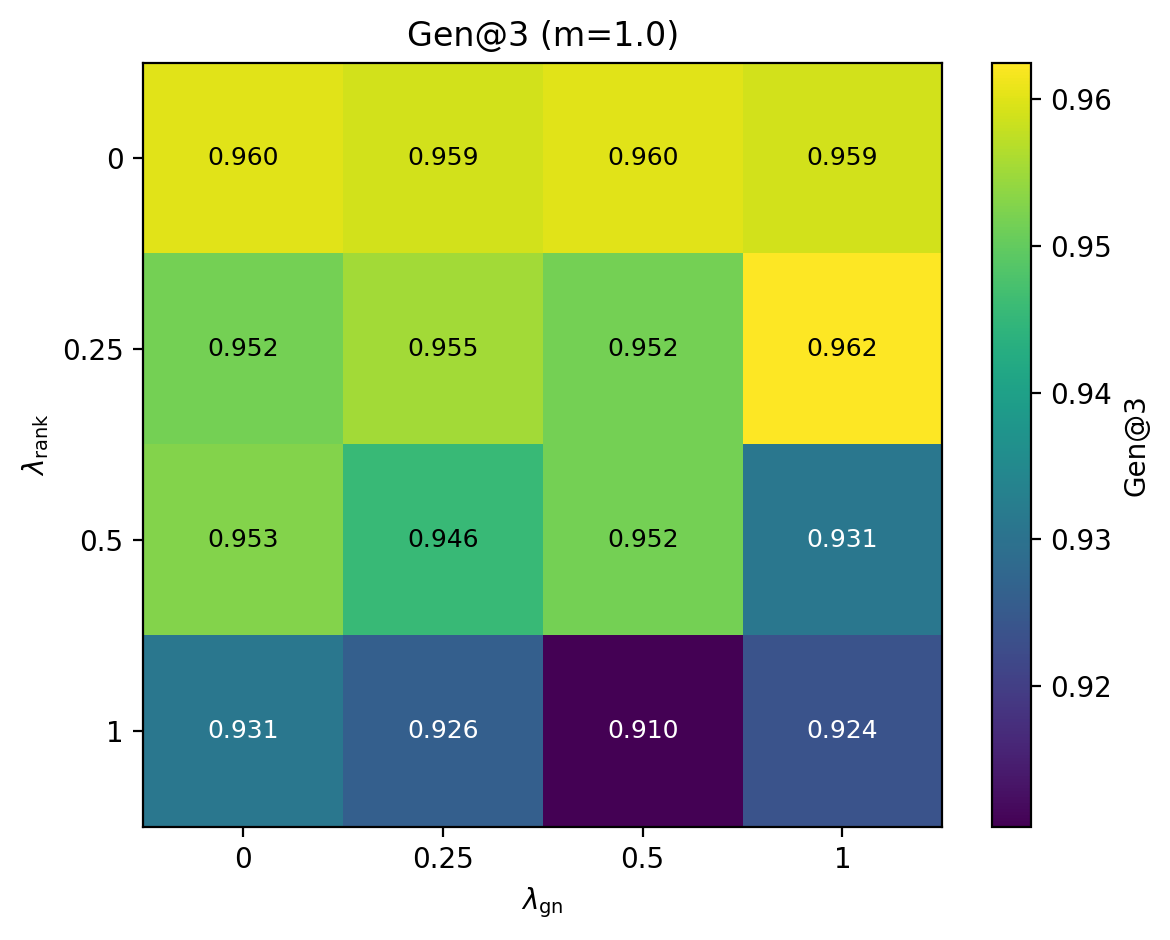}
\includegraphics[width=0.3\linewidth]{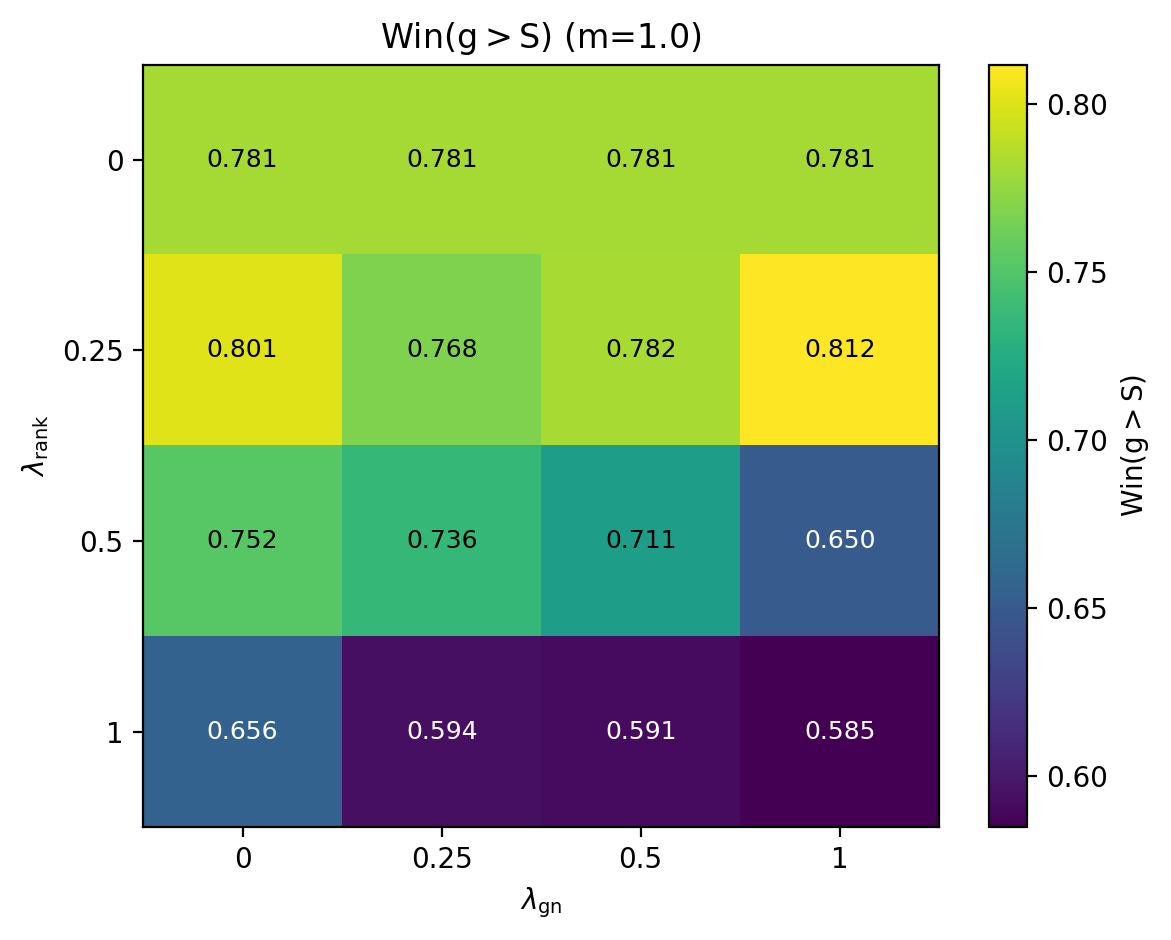}
\includegraphics[width=0.3\linewidth]{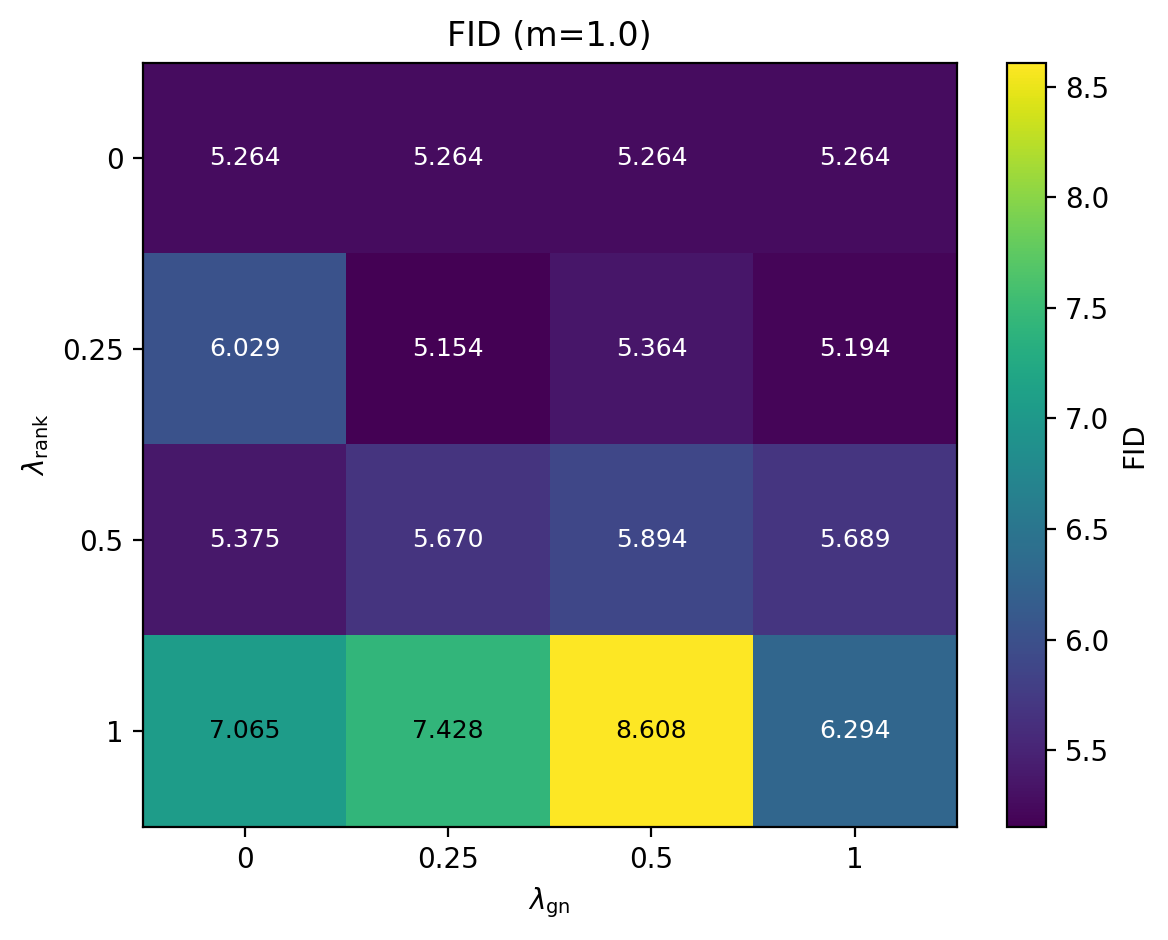}
\\
\includegraphics[width=0.3\linewidth]{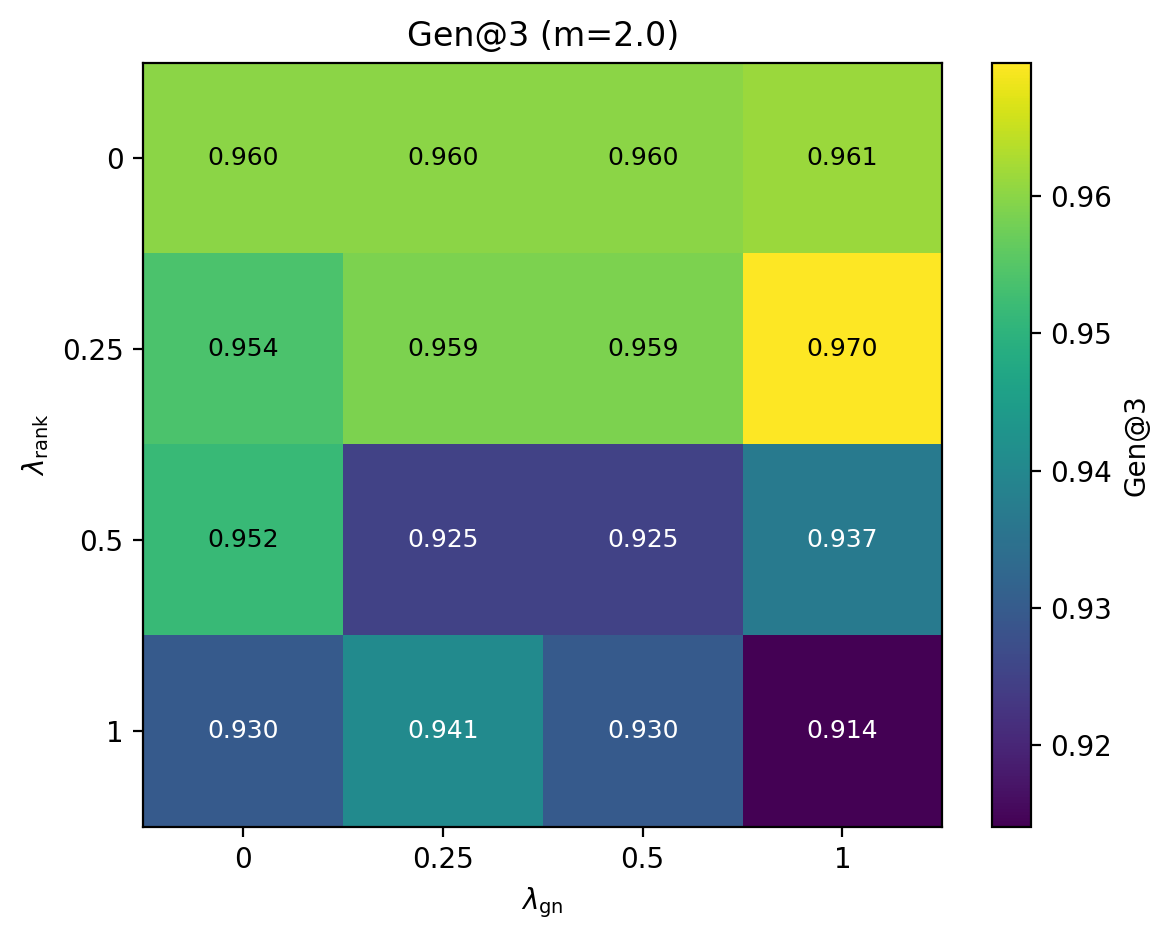}
\includegraphics[width=0.3\linewidth]{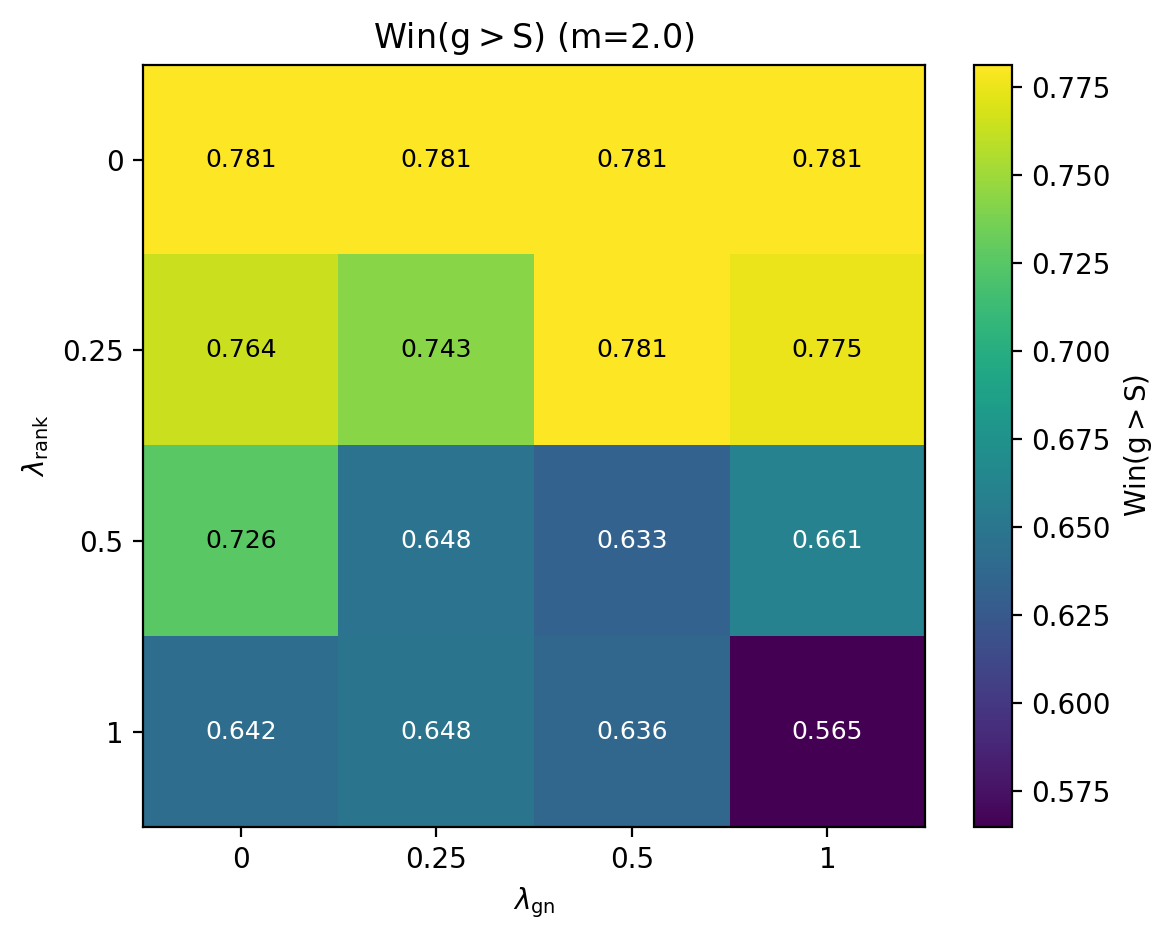}
\includegraphics[width=0.3\linewidth]{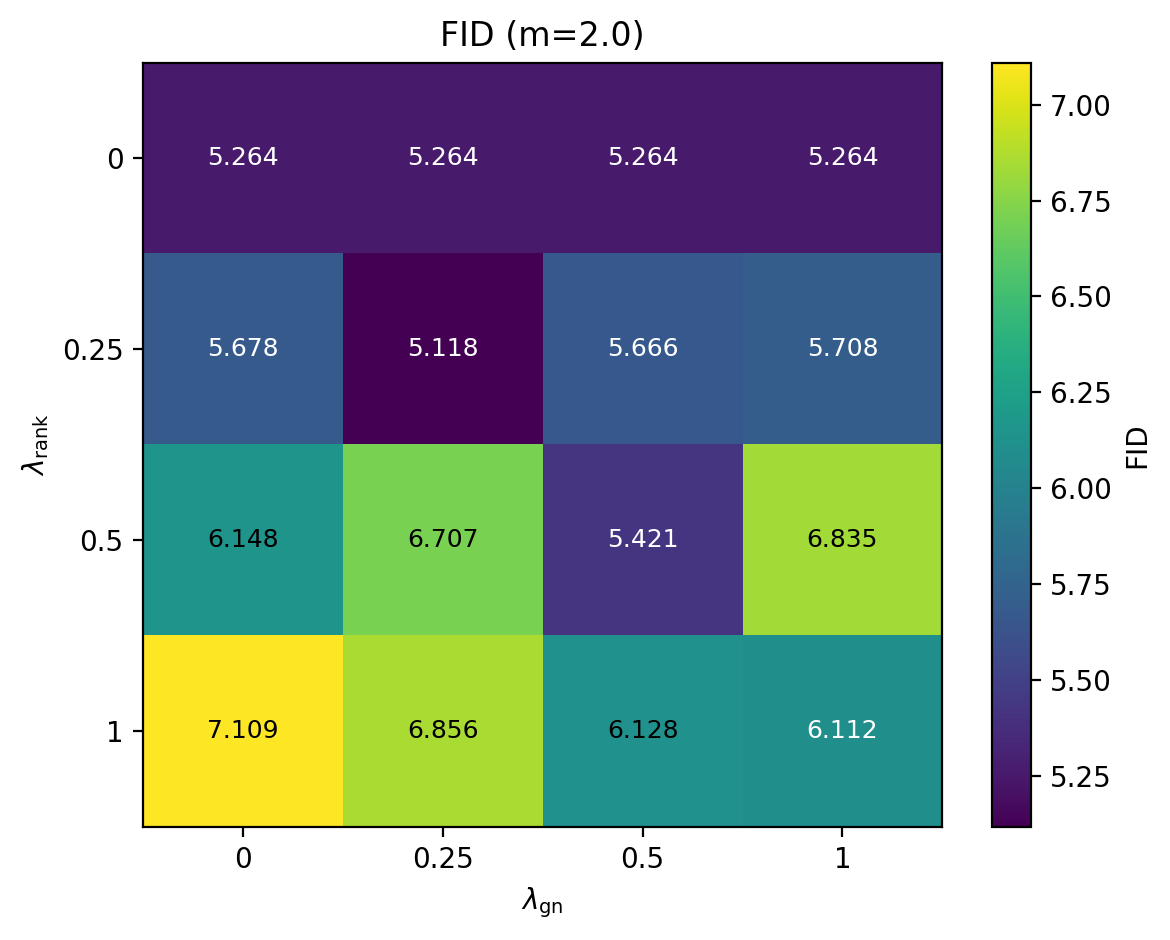}
\\
\caption{Hyperparameter sensitivity heatmaps under different ranking margins. We show Gen@3, Win(g$>$S), and FID as functions of $\lambda_{\mathrm{rank}}$ and $\lambda_{\mathrm{gn}}$.}
\label{fig:supp_hparam_heatmaps}
\end{figure*}

\begin{table}[t]
\centering
\caption{Representative hyperparameter configurations selected from the full sweep. We emphasize Gen@3, Win(g$>$S), and Win(g$>$G), together with FID and Diversity.}
\label{tab:supp_hparam_summary}
\resizebox{\linewidth}{!}{%
\begin{tabular}{lccccccccc}
\toprule
Config & $m$ & $\lambda_{\mathrm{rank}}$ & $\lambda_{\mathrm{gn}}$ & {Win(g$>$N)}$\uparrow$ & {Win(g$>$S)}$\uparrow$ & {Win(g$>$G)}$\uparrow$ & Gen@3$\uparrow$ & FID$\downarrow$ & Diversity$\uparrow$ \\
\midrule
C1 & 2.00 & 0.25 & 1.00 & 0.9939 & 0.7748 & 0.2785 & 0.9697 & 5.7083 & 4.1561 \\
C2 & 0.50 & 0.50 & 0.00 & 0.9964 & 0.8287 & 0.3057 & 0.9625 & 5.8396 & 4.1884 \\
C3 & 0.00 & 0.50 & 0.00 & 0.9927 & 0.7760 & 0.2548 & 0.9443 & 4.6552 & 4.7315 \\
C4 & 1.00 & 0.00 & 0.25 & 0.9976 & 0.7809 & 0.2609 & 0.9588 & 5.2638 & 5.3005 \\
\bottomrule
\end{tabular}
}
\end{table}

\subsection{Inference Efficiency}
\label{Sec:inference}
\begin{table}[ht]
\centering
\caption{Inference Efficiency on a single NVIDIA A100 (80GB).}
\resizebox{0.5\linewidth}{!}{
\begin{tabular}{lc}
\toprule
\textbf{Metric} & \textbf{Value} \\
\midrule
Token generation speed & 63.6 tokens/s \\
Motion generation speed & 1.74 turns/s \\
End-to-end generation speed & 1.66 turns/s \\
Average latency per sample & $\sim$0.60 s \\
VQ-VAE decoding speed & 39.12 turns/s \\
\bottomrule
\end{tabular}
}
\label{tab:time}
\end{table}

Table \ref{tab:time} lists the inference efficiency of the proposed ReactMotion.  During inference, ReactMotion runs on a single NVIDIA A100 80GB GPU and autoregressively generates listener motion tokens conditioned on the speaker’s multimodal inputs. 
In our evaluation, the model generates 50 listener reactive motions corresponding to 50 speaker utterances. In total, it produces 1,830 motion tokens in 28.8 seconds, achieving a generation throughput of 63.6 tokens per second and 1.74 motion sequences per second, which corresponds to an average latency of approximately 0.60 seconds per listener motion sequence.

The generated motion tokens are then decoded into joint sequences using the VQ-VAE decoder. The decoder processes 39.1 motion sequences per second, introducing minimal computational overhead. As a result, the complete pipeline achieves an end-to-end throughput of 1.66 motion sequences per second.
These results indicate that ReactMotion maintains a favorable balance between model capacity and inference efficiency, enabling near real-time reactive motion generation in conversational scenarios.

\subsection{More Details of User Study}
\label{sec:supp_user_study}

We conducted a user study on the Tencent Questionnaire platform to evaluate the listener motions generated by \textbf{ReactMotion} (\textbf{Ours}) against two generative baselines, namely the \textbf{CE} variant and \textbf{LLM}$\rightarrow$\textbf{MG-MotionLLM}~\textsuperscript{*}, as well as the \emph{best-in-group} \textbf{Silver} reference. 
A total of 59 volunteers (16 female and 43 male), all with relevant backgrounds in machine learning or deep learning, participated in the study through an online survey. 
In each trial, participants were presented with a pair of listener-motion videos (A/B) conditioned on the same speaker utterance, with the speaker’s transcript displayed and the corresponding audio played. 
They were asked to choose which video exhibited the more appropriate reactive listener motion. 
To avoid positional bias, the two compared motions were randomly assigned to the A/B positions. 
Each participant completed 36 trials, covering six speaker utterances with six pairwise comparisons per condition. 
For the \textbf{Silver} condition, we selected the best candidate within each speaker-condition group based on its motion caption and rendered motion clip. 

The results in Fig.~\ref{fig:user_study} reveal three notable findings.
First, \textbf{Ours} is consistently preferred over both generative baselines, achieving 67.8\% preference against \textbf{CE} and 72.0\% against \textbf{LLM}$\rightarrow$\textbf{MG-MotionLLM}, which demonstrates the advantage of our unified multimodal Seq2Seq formulation over both standard CE training and cascaded generation pipelines.
Second, although the \textbf{Silver} reference remains stronger overall, \textbf{Ours} is substantially closer to \textbf{Silver} than either baseline: \textbf{Ours} receives 44.1\% of the votes against \textbf{Silver}, whereas \textbf{CE} and \textbf{LLM}$\rightarrow$\textbf{MG-MotionLLM} receive only 31.9\% and 31.4\%, respectively.
This indicates that the motions generated by \textbf{Ours} are perceptually much closer to high-quality in-group references.
Third, these results highlight the effectiveness of the proposed group-wise preference learning objective, which explicitly models the ordering among Gold, Silver, and Negative reactions and leads to more appropriate listener behaviors under human evaluation.
At the same time, the remaining gap between \textbf{Ours} and \textbf{Silver} suggests that reactive listener motion generation remains challenging, leaving room for further improvement in motion naturalness, contextual precision, and diversity.

\subsection{Failure Cases}
\label{sec:supp_failure}

While the model effectively generates contextually appropriate listener motions in many scenarios, capturing deeper conversational intent in complex dialogues remains challenging. In ambiguous or long-tail situations where appropriate listener behavior requires deeper intent understanding, the current model may still exhibit limited robustness. This highlights a promising research direction for future work to further enhance intent-aware interaction modeling in dyadic interaction.

\section{Limitations}
\label{sec:supp_limitations}

Since we are the first to explore this task, we design a relatively simple yet effective model architecture to maintain training stability and computational efficiency. This design allows us to validate the core idea of our approach without introducing excessive architectural complexity. The proposed approach already achieves promising results, demonstrating its feasibility and effectiveness. Nevertheless, there remains a large potential for further improvement. Future work could explore more advanced network architectures and more sophisticated training techniques to further enhance performance.

\bibliographystyle{splncs04}
\bibliography{main}
\end{document}